\documentclass[pdflatex,sn-mathphys]{sn-jnl}% Math and Physical Sciences Numbered Reference Style
\usepackage{array}
\usepackage[table]{xcolor}
\usepackage{colortbl}
\usepackage[authoryear]{natbib}
\usepackage{graphicx}%
\usepackage{makecell}
\usepackage{multirow}%
\usepackage{amsmath,amssymb,amsfonts}%
\usepackage{amsthm}%
\usepackage{mathrsfs}%
\usepackage[title]{appendix}%
\usepackage{float}
\usepackage{textcomp}%

\usepackage{manyfoot}%
\usepackage{booktabs}%
\usepackage{algorithm}%
\usepackage{algorithmicx}%
\usepackage{algpseudocode}%
\usepackage{listings}%
\usepackage[table]{xcolor}
\usepackage{colortbl}
\usepackage{fp}
\usepackage{numprint} % Optional, for nice number formatting

\definecolor{lightgreen1}{RGB}{230, 245, 230}
\definecolor{lightgreen2}{RGB}{200, 230, 200}
\definecolor{lightgreen3}{RGB}{170, 215, 170}
\definecolor{lightgreen4}{RGB}{140, 200, 140}
\definecolor{truthblue}{RGB}{50, 100, 180}
\definecolor{green1}{RGB}{235, 245, 235}  % lightest
\definecolor{green2}{RGB}{210, 235, 210}
\definecolor{green3}{RGB}{180, 220, 180}
\definecolor{green4}{RGB}{150, 200, 150}  
\definecolor{blue1}{RGB}{235, 245, 255}  
\definecolor{blue2}{RGB}{210, 230, 245}  
\definecolor{blue3}{RGB}{180, 210, 235}  
\definecolor{blue4}{RGB}{150, 185, 220}
\definecolor{teal1}{RGB}{240, 255, 250}
\definecolor{teal2}{RGB}{221, 247, 238}
\definecolor{teal3}{RGB}{202, 239, 226}
\definecolor{teal4}{RGB}{183, 231, 214}
\definecolor{teal5}{RGB}{164, 223, 202}
\definecolor{teal6}{RGB}{145, 215, 190}
\definecolor{teal7}{RGB}{126, 207, 178}
\definecolor{teal8}{RGB}{107, 199, 166}
\definecolor{teal9}{RGB}{88, 191, 154}

\definecolor{brown1}{RGB}{255, 242, 220}
\definecolor{brown2}{RGB}{245, 231, 206}
\definecolor{brown3}{RGB}{235, 220, 192}
\definecolor{brown4}{RGB}{225, 209, 178}
\definecolor{brown5}{RGB}{215, 198, 164}
\definecolor{brown6}{RGB}{205, 187, 150}
\definecolor{brown7}{RGB}{195, 176, 136}
\definecolor{brown8}{RGB}{185, 165, 122}
\definecolor{brown9}{RGB}{175, 154, 108}
\definecolor{clustering}{HTML}{E0EBFF}%{E6F2FF}
\definecolor{projection}{HTML}{FFE7E0}%{FFE6E6}
\definecolor{degree}{HTML}{F0EBFF}%{E6FFE6}
\theoremstyle{thmstyleone}%
\theoremstyle{thmstyletwo}%

\theoremstyle{thmstylethree}%

\begin{document}

\title[Article Title]{Understanding the Effect of Knowledge Graph Extraction Error on Downstream Graph Analyses: A Case Study on Affiliation Graphs}


\author*[1]{\fnm{Erica} \sur{Cai}}\email{ecai@cs.umass.edu}

\author*[1]{\fnm{Brendan} \sur{O'Connor}}\email{brenocon@cs.umass.edu}

\affil*[1]{\orgdiv{College of Information and Computer Science}, \orgname{University of Massachusetts Amherst}}%, \orgaddress{\street{Street}, \city{City}, \postcode{100190}, \state{State}, \country{Country}}}

%\affil[2]{\orgdiv{Department}, \orgname{Organization}, \orgaddress{\street{Street}, \city{City}, \postcode{10587}, \state{State}, \country{Country}}}

%\affil[3]{\orgdiv{Department}, \orgname{Organization}, \orgaddress{\street{Street}, \city{City}, \postcode{610101}, \state{State}, \country{Country}}}


\abstract{
Knowledge graphs (KGs) are useful for analyzing social structures, community dynamics, institutional memberships, and other complex relationships across domains from sociology to public health. While recent advances in large language models (LLMs) have improved the scalability and accessibility of automated KG extraction from large text corpora, the impacts of extraction errors on downstream analyses are poorly understood, especially for applied scientists who depend on accurate KGs for real-world insights. To address this gap, we conducted the first evaluation of KG extraction performance at two levels: (1) micro-level edge accuracy, which is consistent with standard NLP evaluations, and manual identification of common error sources; (2) macro-level graph metrics that assess structural properties such as community detection and connectivity, which are relevant to real-world applications. Focusing on affiliation graphs of person membership in organizations extracted from social register books, our study identifies a range of extraction performance where biases across most downstream graph analysis metrics are near zero. However, as extraction performance declines, we find that many metrics exhibit increasingly pronounced biases, with each metric tending toward a consistent direction of either over- or under-estimation. Through simulations, we further show that error models commonly used in the literature do not capture these bias patterns, indicating the need for more realistic error models for KG extraction. Our findings provide actionable insights for practitioners and underscores the importance of advancing extraction methods and error modeling to ensure reliable and meaningful downstream analyses.}

\keywords{knowledge graph extraction, affiliation graph, bipartite graph, error analysis, error propagation, graph analysis}



\maketitle

\section{Introduction}\label{sec1}

Knowledge graphs (KGs)---also known as networks, where nodes represent entities and edges represent relationships between those entities---are often extracted from text to support research and applications across healthcare, social sciences, law enforcement, education, chemistry, and more. In healthcare, KGs from clinical records and article abstracts \citep{Chandak2023-ps,Hong2021-oe,xu2020building} enable cancer analytics and diagnosis \citep{hasan2020knowledge,Li2023-ug}, and facilitate deeper understanding of diabetes complications \citep{Wang2020-dw} and rare diseases \citep{Zhu2020-zn}. In education, KGs extracted from text can help to organize subject content \citep{su2020automatic,chen2018knowedu}. Law enforcement can use KGs from interview transcripts to guide investigations \citep{pandey2020building}, while historians have extracted KGs from documents to reveal Renaissance political and elite relationships \citep{Padgett1993RobustAA}. Sociologists and others have extracted KGs from text to study spatial-social ties \citep{Small2019TheRO}, narrative sequences \citep{BEARMAN200069}, policymaker networks \citep{heclo1978}, and nonfatal gunshot injury patterns \citep{Papachristos2014-lk}. KG extraction from structured citation data and digitized texts also helps with efforts of measuring article impact and of testing sociological theories respectively \citep{bollen2009,latour2012}. These downstream analyses on KGs---often considered more important than extraction itself---provide insights through metrics related to centrality (e.g., to find key actors \citep{freeman1978,zhou2014human,hafnerburton2006}), community detection (e.g., identifying gangs or stakeholder coalitions \citep{sparrow1991,prell2009}), and shortest path analysis (e.g., tracing goods in supply chains \citep{nagurney2010}), among others. 

Both real-world applications and the NLP literature are abundant with research that has addressed the task of extracting KGs, but until recently, their approaches have not had much overlap. Historically, KG extraction in practical applications relied heavily on manual, time-intensive methods \citep{GUGLER2003625,clougherty2014cross,carron2013sagas,edwards2020one} or simpler automated techniques that lacked robust evaluation in the domain \citep{DRURY2022social}. Meanwhile, thousands of papers in the NLP literature aim to improve automated KG extraction, but most methods require substantial amounts of training examples to achieve strong performance, and these methods have typically addressed the more constrained task of extracting relationships between entities that are close to each other in the text, either in the same sentence or paragraph. These constraints have limited the adoption of NLP advances in real-world KG extraction efforts. However, recent advancements in large language models (LLMs) are now bridging this gap by enabling scalable KG extraction with minimal training data, efficient performance over vast amounts of text, and extraction of entity relationships across longer spans of text. Further, LLMs have demonstrated promising performance in domain-specific KG extraction from scientific \citep{dagdelen2024}, legal \citep{li2024construction}, and biomedical texts \citep{jimenez-gutierrez-etal-2022-thinking}, further aligning NLP research with real-world applications.

Given the valuable insights that can be gained from KG analyses, and the promise of recent automated extraction methods requiring small amounts of training data, understanding how extraction errors may bias downstream analyses and lead to flawed conclusions in real-world applications becomes increasingly important to practitioners. The NLP literature, which has introduced hundreds of KG extraction methods and provides extensive resources for their evaluation, seems to be the most relevant reference point for addressing this issue. However, the NLP literature does not provide or use ground truth datasets that pair texts with large, connected knowledge graphs, making investigation of how extraction errors affect downstream analyses difficult. Existing datasets often label relationships between unrelated entities that do not form a connected KG or label small KGs corresponding to isolated paragraphs of text. Even when labeled entities and relationships correspond to larger, connected KGs, the performance of state-of-the-art models remains below 50 F1, which is too low to derive meaningful real-world insights. Therefore, current research in the NLP literature does not provide adequate resources for addressing social measurement challenges effectively.

The lack of sufficient ground truth data to evaluate the effect of KG extraction errors on downstream graph analyses, yet persistent interest in this effect within the network science literature, has led many studies to rely on simulated errors instead. However, many of these simulations are limited. Most only introduce simple random node or edge perturbations, which do not capture the complexity of errors in real-world KG extraction from text. Additionally, simulations are typically performed on synthetic graphs (e.g., Erdős-Rényi models) or on networks derived from structured tabular data, rather than on graphs extracted from free-form text. Moreover, most simulations focus only on a narrow set of graph metrics, such as node centrality, while neglecting other important measures such as community detection and degree distributions that are useful for practical applications.

To our knowledge, no prior work has jointly assessed KG extraction errors at the edge level, which is the standard evaluation in hundreds of NLP papers, and analyzed how these errors affect downstream graph analyses that are relevant to real-world applications. To address this gap, we use three new datasets (a subset of \textsc{AffilKG}; \citet{cai2025relationextractionentirebooks}) of real bipartite affiliation graphs extracted from complete scanned historical social register books, in which edges represent membership relationships between individuals and organizations. Affiliation graphs are widely used to model memberships in elite groups, terrorist networks, and diverse social contexts \citep{corradi2024,isah2015bipartite,holme2007korean,Marotta2014BankFirmCN,duxburyidentifying}, and insights from these analyses are broadly generalizable to most bipartite graphs, as these graphs are all characterized by a single relationship type connecting two kinds of nodes. Analyses of bipartite networks yield valuable findings: degree counts for patient–agency networks have informed healthcare expenditure models \citep{omalley2023diff}; artist-show clustering has had strong correlation with  artistic success \citep{uzzi2005collab}; and projections from homicide investigation networks---connecting people and locations---enable construction of person-to-person co-location graphs for suspect analysis \citep{bichler2017tactical}. By focusing our study on this widely applicable class of KGs, our findings are both specific enough for rigorous evaluation and broadly relevant to practitioners in various domains.

\begin{figure*}
    \centering
    \includegraphics[scale=0.35]{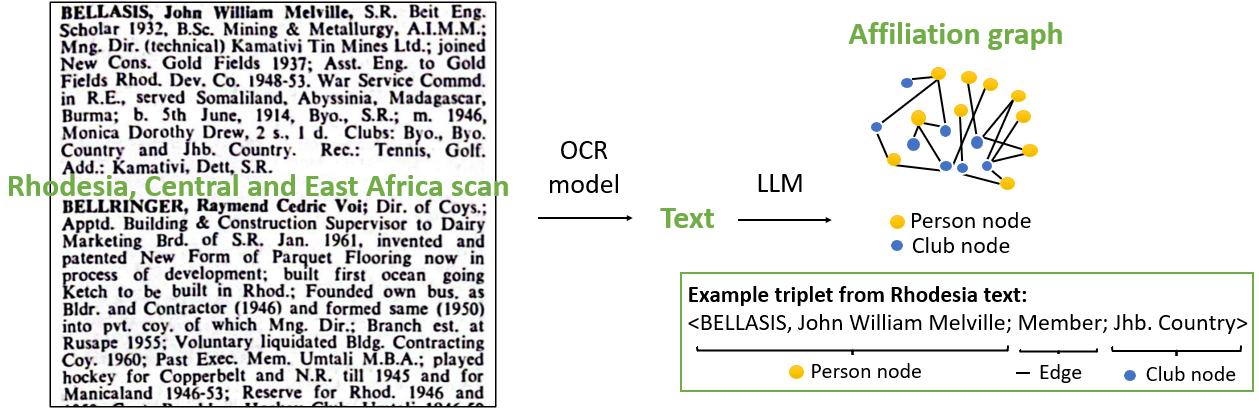}
    %\vspace{-.2em}
    \caption{Pipeline of converting scans of historical text to affiliation graphs, where highlighted text indicates clubs.
    % on extracting articles about randomly selected disaster events of different classes.
    }
    \vspace{-.4em}
    \label{fig:pipeline}
\end{figure*}

By conducting both a micro-analysis---assessing edge-level performance of automated KG extraction on $\langle \textsc{Person},\textsc{Member},\textsc{Club}\rangle$ tuples---and a macro-analysis, which examines how extraction errors affect downstream graph analyses (e.g., degree distributions, centrality, clustering, and co-membership projections) on \textsc{AffilKG} datasets, we find and reveal several valuable insights for practitioners. We observe a general trend of lower precision than recall in relationship extraction, and manual inspection of errors uncovers consistent patterns, such as spelling mistakes, conflation of distinct nodes, and misidentification of nearby text entities as nodes. These types of errors result in extracted graphs with a consistently larger number of nodes compared to the ground truth graphs across most KG extraction methods and datasets. At the level of downstream analysis, we find that extracted graphs exhibit consistent patterns of positive or negative biases relative to ground truth, with the magnitude and direction of these biases varying according to the type of analysis performed. However, when extraction achieves a certain high range of F1 performance, the distortion in most downstream analyses between extracted and ground truth networks is near zero. We hope these findings help practitioners better understand the effect of KG extraction performance on downstream analyses, informing the use of automated methods and enabling the extraction of meaningful insights from large-scale, real-world applications.

Finally, to reduce the need for labor-intensive labeling of ground truth KGs in future studies, we explored whether simulated error could serve as a proxy for understanding the effect of real errors on downstream analyses. We compared the effects of empirically observed errors with those produced by common used error models from the network science and applied literature, as well as with simple error models designed to reflect error patterns we identified through manual inspection. However, our findings show that the most common models were not able to replicate even the general trend of bias direction observed in empirically-extracted KGs. Further, while models that were tailored to better match the observed types of error produced effects that were more consistent with the observed bias direction in empirically-extracted KGs, they consistently overestimated the amount of raw bias by at least four times across all metrics. The inability of simple simulated errors, as modeled in existing approaches, to replicate real-world error patterns underscores the need for more realistic error models that capture the complexity of KG extraction methods for text.

\vspace{0.3em}
\noindent In summary, we contribute the following:

\begin{itemize}
\item \textbf{Micro-analysis}: An empirical evaluation of real-world KG extraction errors at the edge level in affiliation networks using standard NLP metrics, and manual error inspection to identify common fine-grained extraction challenges.
\item \textbf{Macro-analysis}: An empirical investigation into how extraction errors affect high-level graph analyses over bipartite affiliation networks, such as centrality, degree distribution, and clustering.
\item \textbf{Simulation}: A comparative study contrasting downstream analytic results using simulated error models versus real extraction errors, revealing limitations of current error models.
\end{itemize}

\section{Datasets: Paired Books and Affiliation Graphs}
\label{sec:data}

Our analysis of how KG extraction errors affect downstream graph analyses relies on \textsc{AffilKG} \citep{cai2025relationextractionentirebooks}, which is a collection of datasets that pair scanned books, along with their text extracted via optical character recognition (OCR), with large, labeled affiliation graphs. In affiliation graphs, edges represent a single relation type (\textsc{member}), connecting two types of nodes: \textsc{person} and \textsc{club}. These graphs are critical for studying community dynamics \citep{PREMSANKAR2015exploratory, tisch2023top} and other phenomena (see Sec~\ref{sec:relwork}). To our knowledge, \textsc{AffilKG} is the only dataset that pairs text with KGs where existing methods can achieve F1 scores above 50, enabling more meaningful insights for real-world applications.

\begin{table}
{\begin{tabular}{lccc}
 Loc./Year & \begin{tabular}{l} 
\# \\
pages
\end{tabular} & \begin{tabular}{l} 
\#~nodes \\
(org / person)
\end{tabular} & \begin{tabular}{l} 
\# \\
edges
\end{tabular} \\
\midrule Denver/1931 & 34 & 248 / 358 & 1352 \\
S. US./1951 & 826&3107 / 6188&13262\\
{\fontsize{8pt}{8pt}\selectfont \hspace{0.6em} median: Tenn.} & {\fontsize{7.6pt}{8pt}\selectfont 47}& {\fontsize{7.6pt}{8pt}\selectfont 174 / 399} & {\fontsize{7.6pt}{8pt}\selectfont 872} \\ 
Rhod./1963 & 169& 767 / 898 & 2367 \\
\hline
\end{tabular}
}

\caption{Details about Denver, Southern US states, and Rhodesia text and affiliation graph pairs.}
\label{tab:metadata-more}
\end{table}

\vspace{0.3em}
\noindent \textbf{Books.} The texts are social register books documenting elite individuals in the Southern United States \citep{whitmarsh1951}, Denver \citep{denver1931}, and Rhodesia \citep{whoswho1963}, each providing a comprehensive record of a specific social network at a given point in time. Widely used by social scientists to study elite populations and their affiliations \citep{obrien2024family,Broad2020social,Baltzell1958philadelphia,obrien2025gender,Broad1996social}, these registers link individuals to clubs, educational institutions, and other organizations, providing valuable insights into elite community structure. As with many high-quality sources on social networks, these books were difficult to access—existing only in print, requiring OCR for digitization, and available primarily through niche sellers.

\vspace{0.3em}
\noindent \textbf{Affiliation Graphs.} The affiliation graphs are bipartite, encoding tuples of the form $\langle$\textsc{person}; \textsc{member}; \textsc{club}$\rangle$, where \textsc{person} and \textsc{club} are the two node types and \textsc{member} is the single edge type \citep{breiger1947duality,lattanzi2009affiliation}. An example tuple is $\langle$\textsc{BELLASIS, John William Melville}; \textsc{Member}; \textsc{Byo.}$\rangle$ from the \textit{Rhodesia} scan (Fig.~\ref{fig:pipeline}). Creating high-quality ground truth labels for such graphs from scanned books is labor-intensive, but \textsc{AffilKG} provides a rare resource with comprehensive, validated annotations for rigorous empirical analysis.

Our work is the first to combine both micro- and macro-analyses of KG extraction from text. At the micro level, we use common NLP evaluation metrics---precision, recall, and F1 scores---for edge tuples. At the macro level, we examine how extraction errors affect downstream graph analyses in domain-specific applications. To bridge different terminologies between NLP and applied domains, we introduce a unified theoretical framework: we define an \textbf{affiliation graph} as a bipartite graph \[G=(V^{indiv}, V^{club}, E)\]
where $V^{indiv}$ and $V^{club}$ are sets of individual and club nodes, respectively, and $E \subseteq V^{indiv} \times V^{club}$ is the edge set. The full vertex set is $V=V^{indiv} \cup V^{club}$.

\section{Micro-level Analysis: Extraction Performance and Error Characterization}
\label{sec:micro}

We first investigate KG extraction errors at a fine-grained structural level, using metrics of precision and recall of edge tuples $\langle node_1, relation, node_2\rangle$. This offers practitioners interpretable insights into how extraction errors affect the low-level structure of KGs, complementing the higher-level analyses discussed in Section~\ref{sec:macro} that are more directly relevant for real-world applications.

\vspace{0.3em}
\noindent \textbf{Evaluating extraction performance with widely-used metrics.} Thousands of papers in the NLP literature evaluate KG extraction methods exclusively using precision, recall, and F1 scores for edge tuples, where each tuple $\langle$\textsc{person}, \textsc{member}, \textsc{club}$\rangle$ uniquely defines an edge in the affiliation graph. A tuple is a true positive only if all three components match the ground truth (see \S\ref{app:eval} for details). %, with some flexibility for name variations (e.g., omitting suffixes of Jr., Sr., II) and additional \textsc{club} details. 
\textbf{Precision} is the fraction of true positive tuples among those extracted: \[Precision = \frac{|E\cap\hat{E}|}{|\hat{E}|}\] where $|\hat{E}|$ is the set of predicted edges. 

\noindent\textbf{Recall} is the fraction of true positives among ground truth tuples: \[Recall = \frac{|E\cap\hat{E}|}{|E|}\] and the \textbf{F1 score} is the harmonic mean of precision and recall: \[F1=\frac{2 \cdot Precision \cdot Recall}{Precision+Recall}\]

\vspace{0.3em}
\noindent \textbf{Experimental setup: OCR and LLM combinations.} 
Our pipeline consists of two steps: (1) converting book scans to machine-readable text via optical character recognition (OCR), and (2) extracting KGs from the text using large language models (LLMs). We compared four OCR models: Google DocumentAI\footnote{\url{https://cloud.google.com/document-ai}} and AWS TextRact\footnote{\url{https://aws.amazon.com/textract/}} , which outperform prior methods \citep{Hegghammer2022ocr}, and Gemini \citep{geminiteam2024geminifamilyhighlycapable} and Claude Sonnet, which also demonstrate strong performance. For KG extraction, we tested seven LLMs: Gemini, Claude Haiku/Sonnet, GPT-4o/mini \citep{openai2024gpt4ocard}, and Llama 3 (8B, 70B) \citep{grattafiori2024llama3herdmodels}. Each model used 1-shot in-context learning with identical prompts (see \S\ref{app:prompts}) and an example for consistency. Our goal is not to maximize extraction accuracy, but to analyze micro-level performance diversity and its effect on downstream graph analyses. Therefore, variation in scores and imperfect performance is both expected and informative.

\vspace{0.3em}
\noindent \textbf{Graph construction.} After running the full OCR model and LLM experiment pipeline to extract affiliation graphs from the book scans, we extracted 28 graphs corresponding to different combinations of the four OCR models and seven LLMs for each of the 15 social registers. To ensure consistency in mapping entities to graph nodes, we implemented heuristic entity name normalization by removing parenthetical details, standardizing punctuation, and using a dictionary to map abbreviations to full words (e.g., converting `Assn' to `Association'), to account for cases where LLMs may expand abbreviations from the text automatically. We applied the same normalization procedures to the ground truth graphs to enable accurate comparison and evaluation.

\begin{figure}
    \centering
    \includegraphics[scale=0.27]{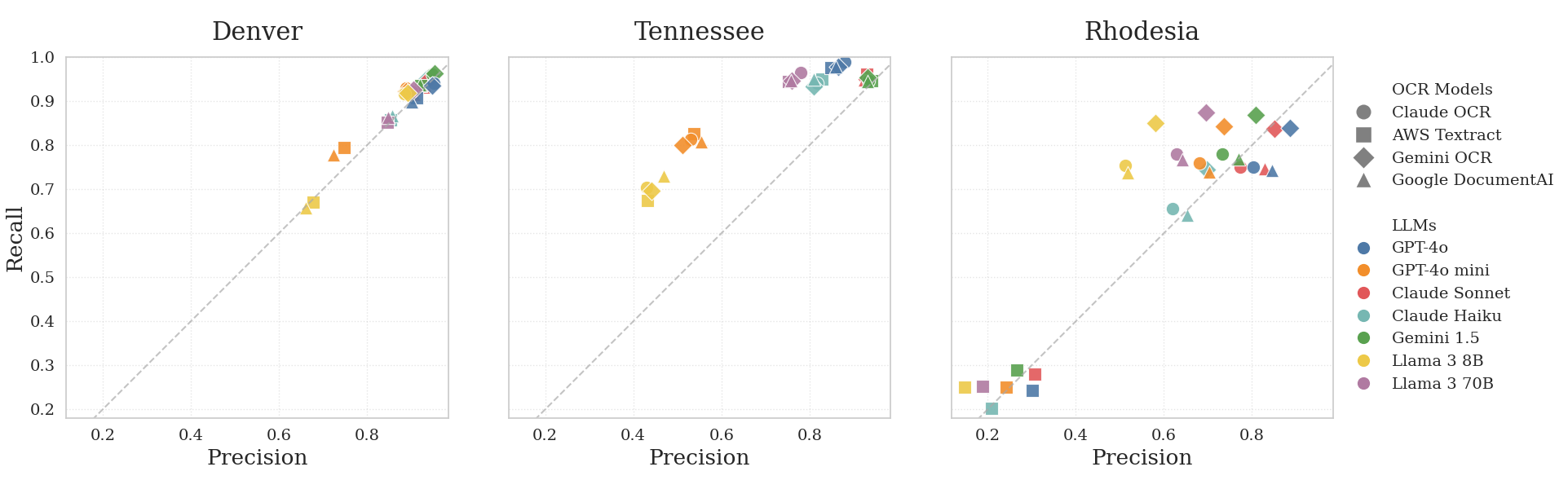}
    \caption{Edge precision (x-axis) and recall (y-axis) for Denver, Tennessee and Rhodesia affiliation graphs of 7 LLMs on 4 versions of OCR-outputted text.
    % on extracting articles about randomly selected disaster events of different classes.
    }
    \label{fig:edge-prec-rec}
\end{figure}

\vspace{0.3em}
\noindent \textbf{Extraction performance across model combinations.} We computed precision, recall, and F1 scores for each combination and visualized these metrics in Fig.~\ref{fig:edge-prec-rec}, which plots precision (x-axis) against recall (y-axis). The plot shows clear patterns: OCR model choice affects performance in \textit{Rhodesia} the most, likely due to differences in handling its two-column text format, while LLM selection strongly affects results in \textit{Tennessee} and across other Southern US states, which exhibit similar trends. Results generally show higher recall than precision across all model combinations, especially for affiliation graphs from southern US states. To understand the nature of these errors better on low-level graph structure, we manually inspect the precision and recall errors.

\vspace{0.3em}
\noindent\textbf{Manual error analysis: categorizing extraction failures.} In each dataset, we randomly sampled 280 false positive edge tuples---ten per OCR and LLM pair---and classified their error types (Table~\ref{tab:error_distribution}). We focused our analysis on false positives, since false negatives are often either random omissions or the result of error patterns already captured by looking at false positives. %For 280 false negative tuples, errors appeared mostly random and were hard to categorize.

\begin{table}[h]
\centering
\small
\renewcommand{\arraystretch}{1.2}
\begin{tabular}{p{7cm} c c c} % Use p{6cm} for the first column to allow wrapping
%%\toprule
 & Denver & Tennessee & Rhodesia \\
\midrule
Minor spelling error of node (within a word) $\to$ new node & 67 & 0 & 0 \\ \arrayrulecolor{lightgray}\hline
Major spelling difference of node (at least one word is incorrect) $\to$ new node & 6 & 26 & 23 \\ \arrayrulecolor{lightgray}\hline
Considered two entities as a single node $\to$ new node & 19 & 0 & 18 \\ \arrayrulecolor{lightgray}\hline
Mistook a different entity in the text (already in the graph) as a node $\to$ redirected edge & 4 & 49 & 28 \\ \arrayrulecolor{lightgray}\hline
Mistook a different entity in the text (not in the graph) as a node $\to$ new node  & 4 &  21 & 29 \\ \arrayrulecolor{lightgray}\hline
Entirely made up a node $\to$ new node & 0 & 4 & 2 \\
\bottomrule
\end{tabular}
\caption{Percent distribution of error types across relations extracted from different texts}
\label{tab:error_distribution}
\end{table}

Our manual inspection of false positive edge tuples revealed common errors such as node spelling mistakes, merging two entities into a single node, and misidentifying nearby entities not present in the graph. Many of these errors introduce new nodes, such as misspelled versions of existing nodes, combinations of distinct names, or entities not present in the ground truth, rather than merely rearranging edges among existing nodes as assumed by many error models in the literature. To assess their structural impact, we examine the raw counts of nodes and edges in the extracted graphs.

\vspace{0.3em}
\noindent\textbf{Structural impact: node and edge count inflation.} The manual error inspection implies that the extracted graphs often contain significantly more nodes than the ground truth graphs. To quantify this discrepancy, we calculated the percentage by which the predicted number of nodes in the extracted graph exceeds the true counts using the formula: $100*(\hat{|V|}-|V|)/|V|$, where $\hat{|V|}$ and $|V|$ are the predicted and true node counts; we applied the same formula to edges ($E$). Table~\ref{tab:nodescomplete} shows results across $7$ LLMs on text outputted by $4$ OCR models which consistently indicate an overestimation of node counts, where discrepancies are often very large. This pattern is less pronounced for edges as shown in Table~\ref{tab:edgescomplete}, but overestimation still occurs. These findings are consistent with Fig.~\ref{fig:edge-prec-rec}, where edge tuple precision tends to be worse than recall.

\begin{table*}
\footnotesize
\centering

\begin{tabular}{llrrrrrrr}

& & Gem-1.5 & Clau Sonn & Gpt-4o & Clau Haiku & 
llama-70b & 4o-mini & llama-8b \\
\midrule

\multirow{3}{*}{TextRact}
 & Denver & \cellcolor{teal2}8.1 & \cellcolor{teal2}5.6 & \cellcolor{teal2}8.4 & \cellcolor{teal2}12.7 & \cellcolor{teal2}13.0 & \cellcolor{teal2}20.0 & \cellcolor{teal2}32.8 \\
 & Tenn.  & \cellcolor{teal2}1.6 & \cellcolor{teal2}1.4 & \cellcolor{teal2}7.3 & \cellcolor{teal2}9.8  & \cellcolor{teal2}19.7 & \cellcolor{teal2}37.5 & \cellcolor{teal2}62.1 \\
 & Rhod.  & \cellcolor{teal2}17.4 & -4.9 & -9.2 & \cellcolor{teal2}6.1 & \cellcolor{teal2}65.6 & \cellcolor{teal2}23.6 & \cellcolor{teal2}122.7 \\
\midrule

\multirow{3}{*}{DocumentAI}
 & Denver & \cellcolor{teal2}7.1 & \cellcolor{teal2}3.0 & \cellcolor{teal2}5.0 & \cellcolor{teal2}13.9 & \cellcolor{teal2}13.0 & \cellcolor{teal2}18.8 & \cellcolor{teal2}32.0 \\
 & Tenn.  & \cellcolor{teal2}1.7 & \cellcolor{teal2}1.2 & \cellcolor{teal2}6.3 & \cellcolor{teal2}13.3 & \cellcolor{teal2}20.1 & \cellcolor{teal2}31.9 & 6\cellcolor{teal2}0.2 \\
 & Rhod.  & \cellcolor{teal2}10.5 & -5.2 & -4.2 & \cellcolor{teal2}8.1 & \cellcolor{teal2}42.8 & \cellcolor{teal2}22.0 & \cellcolor{teal2}78.6 \\
\midrule

\multirow{3}{*}{Claude-3-sonnet}
 & Denver & \cellcolor{teal2}4.5 & \cellcolor{teal2}3.8 & \cellcolor{teal2}3.1 & \cellcolor{teal2}5.9 & \cellcolor{teal2}8.1 & \cellcolor{teal2}12.5 & \cellcolor{teal2}15.7
 \\
 & Tenn.  & \cellcolor{teal2}2.8 & \cellcolor{teal2}1.9 & \cellcolor{teal2}6.6 & \cellcolor{teal2}12.0 & \cellcolor{teal2}18.3 & \cellcolor{teal2}37.2 & \cellcolor{teal2}64.0
\\
 & Rhod.  & \cellcolor{teal2}21.0 & \cellcolor{teal2}5.5 & \cellcolor{teal2}4.5 & \cellcolor{teal2}19.9 & \cellcolor{teal2}51.2 & \cellcolor{teal2}32.7 & \cellcolor{teal2}84.2
 \\
\midrule

\multirow{3}{*}{Gemini-1.5-pro} 
 & Denver & \cellcolor{teal2}3.1 & \cellcolor{teal2}2.6 & \cellcolor{teal2}2.1 & \cellcolor{teal2}5.8 & \cellcolor{teal2}7.1 & \cellcolor{teal2}10.9 & \cellcolor{teal2}14.9
 \\
 & Tenn.  & \cellcolor{teal2}2.4 & \cellcolor{teal2}1.2 & \cellcolor{teal2}5.6 & \cellcolor{teal2}11.7 & \cellcolor{teal2}20.6 & \cellcolor{teal2}40.3 & \cellcolor{teal2}61.6
\\
 & Rhod.  &\cellcolor{teal2}18.1 & \cellcolor{teal2}2.5 & \cellcolor{teal2}2.3 & \cellcolor{teal2}17.1 & \cellcolor{teal2}48.1 & \cellcolor{teal2}32.1 & \cellcolor{teal2}79.5
\\
\bottomrule
\end{tabular}
\caption{Percent by which the predicted number of nodes exceeds the true number of nodes across 7 LLMs on 4 OCR-outputted texts. }
\label{tab:nodescomplete}
\end{table*}

These results suggest that KG extraction errors produce graphs with more false positive edges and significantly more nodes than the ground truth, often due to spelling mistakes, treatment of two distinct entities as a single node, or misidentification of different entities in the text as nodes. Extracted graphs are often larger, sometimes substantially. This inflation contrasts with some existing common error models, which add false positive edges randomly without increasing node counts or focus on specific narrow types of error. In Section~\ref{sec:sim}, we investigate this further, finding that simulation analyses may fail to fully capture the effects of real-world KG extraction errors.%, as explored in Section~\ref{sec:sim}. This disconnect underscores the need for error models that capture the diverse and complex types of error resulting from KG extraction.

\begin{table*}{\footnotesize
\centering

\begin{tabular}{llrrrrrrr}

& & Gem-1.5 & Clau Sonn & Gpt-4o & Clau Haiku & 
llama-70b & 4o-mini & llama-8b \\
\midrule

\multirow{3}{*}{TextRact}
 & Denver & \cellcolor{teal2}1.5 & \cellcolor{teal2}0.3 & -0.4 & \cellcolor{teal2}0.4 & \cellcolor{teal2}0.7 & \cellcolor{teal2}6.4 & -1.2
 \\
 & Tenn.  & \cellcolor{teal2}0.5 & \cellcolor{teal2}3.3 & \cellcolor{teal2}14.9 & \cellcolor{teal2}14.8 & \cellcolor{teal2}25.5 & \cellcolor{teal2}53.8 & \cellcolor{teal2}56.2
 \\
 & Rhod.  & \cellcolor{teal2}8.7 & -9 & -19.3 & -2.5 & \cellcolor{teal2}34.4 & \cellcolor{teal2}3.9 & \cellcolor{teal2}70.3
 \\
\midrule

\multirow{3}{*}{DocumentAI}
 & Denver & \cellcolor{teal2}1.7 & \cellcolor{teal2}0.9 & -0.2 & \cellcolor{teal2}1.2 & \cellcolor{teal2}2 & \cellcolor{teal2}7.6 & -0.4
 \\
 & Tenn.  & \cellcolor{teal2}1.4 & \cellcolor{teal2}2.5 & \cellcolor{teal2}13.8 & \cellcolor{teal2}17.2 & \cellcolor{teal2}24.8 & \cellcolor{teal2}45.6 & \cellcolor{teal2}55.7 \\
 & Rhod.  & -0.1 & -9.8 & -12.2 & -1.9 & \cellcolor{teal2}19.4 & \cellcolor{teal2}5.1 & \cellcolor{teal2}42.2 \\
\midrule

\multirow{3}{*}{Claude-3-sonnet}
 & Denver & \cellcolor{teal2}1.4 & \cellcolor{teal2}1 & -1 & \cellcolor{teal2}1.1 & \cellcolor{teal2}2.1 & \cellcolor{teal2}4.7 & \cellcolor{teal2}3.6

 \\
 & Tenn.  & \cellcolor{teal2}1.4 & \cellcolor{teal2}1.7 & \cellcolor{teal2}12.3 & \cellcolor{teal2}15.4 & \cellcolor{teal2}23.7 & \cellcolor{teal2}53.2 & \cellcolor{teal2}63.2

\\
 & Rhod.  & \cellcolor{teal2}6.6 & -3 & -6.5 & \cellcolor{teal2}5.8 & \cellcolor{teal2}24.1 & \cellcolor{teal2}11.4 & \cellcolor{teal2}47.1

 \\
\midrule

\multirow{3}{*}{Gemini-1.5-pro} 
 & Denver & \cellcolor{teal2}1 & \cellcolor{teal2}1 & -1.3 & \cellcolor{teal2}1.7 & \cellcolor{teal2}2.2 & \cellcolor{teal2}3.8 & \cellcolor{teal2}2.9

 \\
 & Tenn.  & \cellcolor{teal2}2.4 & \cellcolor{teal2}2.3 & \cellcolor{teal2}12.8 & \cellcolor{teal2}15.3 & \cellcolor{teal2}24.7 & \cellcolor{teal2}56.2 & \cellcolor{teal2}57.7

\\
 & Rhod.  &\cellcolor{teal2} 7.4 & -1.6 & -5.3 & \cellcolor{teal2}6.8 & \cellcolor{teal2}25.6 & \cellcolor{teal2}14.5 & \cellcolor{teal2}46.5

\\
\bottomrule
\end{tabular}}
\caption{Percent by which the predicted number of edges exceeds the true number of nodes across 7 LLMs on 4 OCR-outputted texts. }
\label{tab:edgescomplete}
\end{table*}

\section{Downstream Analyses of Bipartite Affiliation Graphs}
\label{sec:relwork}
%\vspace{0.3em}
To provide practitioners with insights into how KG extraction errors affect downstream analyses on real-world networks, we conducted a comprehensive literature survey to identify common downstream graph analyses. These are widely performed and studied in fields such as economics, network science, and biology. In particular, several common types of analyses with broad practical application are degree distributions, clustering, and co-membership projection analyses onto a single node type.

\vspace{0.3em}
\noindent\textbf{\colorbox{degree}{Node-level degree statistics and importance.}} 
In bipartite networks, a node's degree is the number of edges it has to nodes of the other type. This metric is valuable for gathering aggregate statistics, such as average counts and variance, identifying important entities based on their connections \citep{duxburyidentifying}, and supporting further downstream analyses, including regressions that address critical questions in social science. For example, high degree nodes in terrorist group and attack-related event networks reveal the most affected targets and most common attack types, and degree distribution helps to reveal the diversity of locations targeted by each groups (e.g., Boko Haram targets the most locations; Fulani extremists use the widest range of attack methods) \citep{corradi2024}. Degree metrics also help to identify important entities in criminal counterfeiting actor-resource bipartite networks \citep{isah2015bipartite}.

Degree information also enables subsequent downstream analyses; for example, regressions using degree counts in home health agency-patient networks estimate health care expenditures \citep{omalley2023diff}. Degree-like metrics, such as the average bills sponsored per legislator, are important for studying legislative cosponsorship and differentiating these networks from other large-scale social structures \citep{fowler2006leg}.

\vspace{0.3em}
\noindent\textit{\textbf{Metrics.}} In our analyses, designed to benefit practitioners, we focus on metrics that capture degree statistics consistent with prior literature; specifically, the \textbf{mean} and \textbf{standard deviation} of node degrees in each partition, $V^{club}$ and $V^{indiv}$. %, of \textbf{mean} and \textbf{standard deviation} of degrees for nodes in each partition $V^{club}$ and $V^{indiv}$. 
In affiliation graphs, the degree of $v^{club} \in V^{club}$ indicates the number of members in club $v^{club}$ and the degree of $v^{indiv} \in V^{indiv}$ indicates the number of clubs that individual $v^{indiv}$ is a member of. %To capture these degree distributions, we compute aggregate statistics of the \textbf{mean} and \textbf{standard deviation} of degrees for nodes in $V^{club}$ and for nodes in $V^{indiv}$, as commonly used in previous studies. 

In our bipartite network application, which involves node types of people and social clubs, club size is particularly important, playing a critical role in understanding how elite communities socialize and cohere, as it conveys status, facilitates socialization, and fosters the development of shared norms within these communities \citep{Accominotti2018cultural,cousin2014globalizing}. Therefore, we are particularly interested in understanding errors of measuring the size for each club across the network, as well as for each of the ten largest clubs, which wield the most influence in the elite population. To assess these errors, we compute average relative mean absolute error (RMAE), defined as:  \[RMAE=\frac{1}{|V^{club}|} \sum_{v^{club}} \frac{|\hat{deg}({v^{club}})-deg({v^{club}})|}{deg({v^{club}})}\] where $\hat{deg}({v^{club}})$ is the true degree of node $v^{club}$ and $deg(v^{club})$ is the degree of $v^{club}$ in the extracted affiliation graph. %Next, we compute the RMAE for the ten largest clubs in each dataset to evaluate errors in the most influential clubs. Measuring the sizes of important clubs is valuable in applications such as xx. 

\vspace{0.3em}
\noindent\textbf{\colorbox{clustering}{Subgraph-level clustering.}} Previous literature uses various approaches to measure clustering in networks. For example, studies constructing bipartite networks of actors and resources involved in crime have examined "cohesion", "group stability", the "rich-club phenomenon", which indicates that highly connected nodes form dense subgroups, transitivity, and small-world properties such as short path lengths and clustering connectivity \citep{isah2015bipartite}. Community detection, which is the process of identifying groups of nodes that are more densely connected to each other than to the rest of the network, has also been an important analysis \citep{zhang2008clustering,murata2009detecting,zhou2018novel,girvan2002community}. In the banking sector, community detection on bank-firm bipartite networks in Japan across 32 years has revealed layered structures within the credit market, characterized by groups of firms over-represented in specific economic sectors \citep{Marotta2014BankFirmCN}.  Specialized methods have been developed for detecting communities in biological bipartite networks \citep{bhavnani2011how,pesantezcabrera2019,pesantez2016detecting}. Other studies focus on quantifying and measuring social exchange and balance \citep{gallos2012how} or develop theoretical metrics to capture various clustering phenomena \citep{opsahl2010triadic,latapy2008basic,lind2005cycles,zhang2008clustering,smallworlds2004robins}. \citet{uzzi2005collab}, constructing artist-show bipartite networks, perform further downstream analyses to find that high clustering correlates with artistic success while short path lengths correlate with financial success of shows.

\vspace{0.3em}
\noindent\textit{\textbf{Metrics.}} In our analyses, we focus on metrics that are consistent with prior literature to ensure practical relevance. These include \textbf{density}, which measures the proportion of edges to possible edges in the network and is defined as: \[D=\frac{|E|}{|V^{indiv}|*|V^{club}|}\] and \textbf{number of connected components}, which counts the number of disjoint subgraphs in the network \citep{newman2001structure,Castellano_2009statistical}, and where a large number suggests the presence of many isolated communities. We also consider the \textbf{number of communities}, defined as densely connected groups identified using algorithms such as greedy modularity \citep{newman2011networks,clauset2004,reichardt2006,newman2004analysis}.

Given that the real affiliation networks typically consist of one large connected component, we additionally measure the \textbf{proportion of nodes in the largest connected component} relative to the total number of nodes, as well as the \textbf{average size of remaining connected components}. These metrics provide insight into the overall connectivity of the network, distinguishing the main core from smaller isolated groups. Additionally, we compute the \textbf{diameter}, which represents the longest shortest path within a subgraph, and the \textbf{average shortest path length} within the main connected component, both of which have been useful in prior studies \citep{Ugander2011TheAO,bearman2004chains,dodds2003b,isah2015bipartite}.

\vspace{0.3em}
\noindent\textbf{\colorbox{projection}{Analyses over projection networks.}} Bipartite networks, consisting of two node types, are often projected onto one type to form a collapsed network, or \textbf{projection network}, where nodes are connected if they share at least one neighbor in the bipartite network. This approach is supported by \citet{feld1981focused}, who notes that social ties often form among individuals participating in shared activities that are centered around a common focus. Criminologists have used projected affiliation networks to uncover connections between Securities and Exchange Commission (SEC) violators and Fortune 500 CEOs \citep{bichler2015white}, identify working relationships among officers through call-for-service co-attendance (e.g., \citet{young2015diffusion}), and study gang structures (e.g., \citet{grund2015ethnic}). For example, \citet{calderoni2015network} revealed emerging leaders in criminal organizations by projecting person-to-person networks based on meeting participation. In serial homicide investigations, bipartite networks of people and locations have been projected onto person-to-person networks to facilitate analysis of different suspects \citep{bichler2017tactical}. In education, student-course bipartite networks have been collapsed onto students to uncover cliques via comembership networks \citep{holme2007korean}. In other studies of crime, bipartite networks of actors and resources have been used to infer relationships between criminals who share common attributes or resources but may not have direct connections \citep{isah2015bipartite}. From shared bill sponsorships, Fowler \citep{fowler2006leg} study constructed legislator-to-legislator networks, revealing how institutional structures significantly increase network density compared to typical social networks.

\vspace{0.3em}
\noindent\textit{\textbf{Metrics.}} In our analyses, we collapse the affiliation network onto one node type---either people or clubs---to construct two types of networks: a comembership network and a network of organizations that share members. In the comembership network, an edge exists between individuals who share at least one common social club. In the network of organizations sharing members, edges connect organizations that have at least one member in common. %This organizational network is particularly useful for (cite).

For both projections, we compute metrics that are widely used in network analysis in the literature: \textbf{density} and \textbf{average clustering coefficient}. Density measures the proportion of actual edges to possible edges in the network and is defined as: $D=\frac{|E|}{|V|(|V|-1)}$
The \textbf{average clustering coefficient} quantifies the extent that nodes tend to form tightly connected groups, by measuring the fraction of possible triangles present for each node. A triangle is a set of three nodes where each node is connected to the other two, forming a closed triplet. The clustering coefficient for a node $v$ is:  \[c_v=\frac{2T(v)}{deg(v)*(deg(v)-1)}\] where $T(v)$ is the number of triangles through $v$ and $deg(v)$ is the degree of $v$ \citep{saramaki2007,onnela2005}. The average clustering coefficient is calculated by taking the mean of $c_v$ across all nodes in the projection network. These metrics provide a robust framework for evaluating the structural properties of the affiliation graphs and their implications for real-world applications.

\section{Macro-level Analysis: Impact of Extraction Errors on Downstream Graph Metrics}
\label{sec:macro}

Using ground truth and extracted affiliation graphs from \textsc{AffilKG} and our OCR/LLM experiments (Section~\ref{sec:micro}), we analyze how KG extraction errors impact downstream graph analyses relevant to real-world applications (Section~\ref{sec:relwork}). This extends our earlier low-level structural findings to provide actionable insights for practitioners.

\vspace{0.3em}
\noindent \textbf{Relative Bias.} For each metric described in Section~\ref{sec:relwork}, we compute the relative bias, defined as:\[rel\text{ } bias = \frac{\hat{metric}-metric}{metric}\]

\noindent where $\hat{metric}$ is the value on the extracted graph and $metric$ is the ground truth value. This quantifies the extent to which the extracted graph overestimates or underestimates the ground truth value and allows comparison across metrics despite variations in their raw magnitudes.

\vspace{0.3em}
\noindent \textbf{Relative Mean Absolute Error (MAE).} We also calculate the relative mean absolute error:

\[rel\text{ } MAE = \frac{|\hat{metric}-metric|}{metric}\] to capture the magnitude of error, and compare it to bias.

\begin{table}[ht]
\centering

\begin{tabular}{>{\raggedright}p{2.2cm}>{\raggedright}p{2cm}cccc}
%\toprule
 & & \multicolumn{2}{c}{\makecell{Comembership Network\\ (Projection Network $V^{indiv}$)}} & \multicolumn{2}{c}{\makecell{Organizations Sharing\\ Members (Projection Network $V^{club}$)}} \\
\cmidrule(lr){3-4} \cmidrule(lr){5-6}
 & & Density & \makecell{Avg Clustering\\ Coefficient} & Density & \makecell{Avg Clustering\\ Coefficient} \\
\midrule
\multirow{3}{*}{\makecell{Ground\\ Truth}} 
 & Denver & {0.491} & {0.804} & {0.042} & {0.816} \\
 & South US (avg) & {0.110} & {0.763} & {0.028} & {0.564} \\
 & Rhodesia & {0.156} & {0.768} & {0.008} & {0.651} \\
\midrule
\multirow{4}{*}{\makecell{Average Rel.\\ Bias over KGs}} 
 & F1 [0.92, 1.00) & \cellcolor{brown1}{-0.046} & \cellcolor{brown1}{-0.008} & \cellcolor{brown2}{-0.074} & \cellcolor{brown1}{-0.016} \\
 & F1 [0.84, 0.92) & \cellcolor{brown3}{-0.136} & \cellcolor{brown1}{-0.032} & \cellcolor{brown3}{-0.132} & \cellcolor{brown2}{-0.070} \\
 & F1 [0.76, 0.84) & \cellcolor{brown4}{-0.205} & \cellcolor{brown1}{-0.062} & \cellcolor{brown4}{-0.227} & \cellcolor{brown2}{-0.090} \\
 & F1 [0.40, 0.76) & \cellcolor{brown7}{-0.401} & \cellcolor{brown2}{-0.107} & \cellcolor{brown7}{-0.429} & \cellcolor{brown3}{-0.143} \\
\midrule
\multirow{4}{*}{\makecell{Average Rel.\\ MAE over KGs}} 
 & F1 [0.92, 1.00) & \cellcolor{teal1}{0.050} & \cellcolor{teal1}{0.017} & \cellcolor{teal2}{0.081} & \cellcolor{teal1}{0.027} \\
 & F1 [0.84, 0.92) & \cellcolor{teal3}{0.157} & \cellcolor{teal1}{0.039} & \cellcolor{teal3}{0.147} & \cellcolor{teal2}{0.084} \\
 & F1 [0.76, 0.84) & \cellcolor{teal4}{0.250} & \cellcolor{teal2}{0.073} & \cellcolor{teal4}{0.245} & \cellcolor{teal2}{0.110} \\
 & F1 [0.40, 0.76) & \cellcolor{teal7}{0.426} & \cellcolor{teal2}{0.122} & \cellcolor{teal7}{0.431} & \cellcolor{teal3}{0.149} \\
\bottomrule
\end{tabular}
\caption{\textbf{{\colorbox{projection}{Analyses over Projection Networks.}}} Comparison of relative bias of metrics that capture analyses of projection networks, and  ground truth over KGs from the three scanned books. Cell shading intensity corresponds to the magnitude of the value, with darker colors indicating larger magnitudes. Brown is used for negative values, and green for positive values.}
\label{tab:projection-metrics-comparison}
\end{table}

\begin{table}[ht]

\begin{tabular}{l l c c c c c c c}
%\toprule
 & & \multicolumn{3}{c}{\makecell{Full network}} & \multicolumn{3}{c}{\makecell{Largest Conn. Comp.}} & {\makecell{Rest \\Comp.}}\\
\cmidrule(lr){3-5} \cmidrule(lr){6-8} \cmidrule(lr){9-9}
 & & {\makecell{\# Conn.\\ Comp.}} & {\makecell{\# Com- \\munity}} & {\makecell{Bipartite \\Density}} & {\makecell{Prop. of\\ Nodes}} & {\makecell{Avg Short.\\ Path Len.}} & {Diam.} & {Avg Size} \\
\midrule
\multirow{3}{*}{\makecell{Ground\\ Truth}} 
 & Denver & {11} & {36} & {0.015} & {0.947} & {3.671} & {10} & {3.200} \\
 & South US (avg) & {31.385} & {46.615} & {0.012} & {0.867} & {5.294} & {14.231} & {2.692} \\
 & Rhodesia & {74} & {104} & {0.003} & {0.868} & {5.100} & {15} & {3.014} \\
\midrule
\multirow{4}{*}{\makecell{Average \\Rel. Bias \\over KGs}} 
 & F1 [0.92, 1.00) & \cellcolor{teal8}{0.467} & \cellcolor{teal3}{0.126} & \cellcolor{brown1}{-0.051} & \cellcolor{brown1}{-0.022} & \cellcolor{teal1}{0.011} & \cellcolor{teal1}{0.005} & \cellcolor{brown2}{-0.068} \\
 & F1 [0.84, 0.92) & \cellcolor{teal9}{0.716} & \cellcolor{teal5}{0.256} & \cellcolor{brown2}{-0.090} & \cellcolor{brown1}{-0.037} & \cellcolor{teal1}{0.032} & \cellcolor{teal1}{0.051} & \cellcolor{brown1}{-0.055} \\
 & F1 [0.76, 0.84) & \cellcolor{teal9}{1.792} & \cellcolor{teal9}{0.717} & \cellcolor{brown4}{-0.172} & \cellcolor{brown2}{-0.084} & \cellcolor{teal1}{0.043} & \cellcolor{teal1}{0.042} & \cellcolor{brown1}{-0.034} \\
 & F1 [0.40, 0.76) & \cellcolor{teal9}{2.697} & \cellcolor{teal9}{1.337} & \cellcolor{brown6}{-0.345} & \cellcolor{brown3}{-0.167} & \cellcolor{teal2}{0.107} & \cellcolor{teal3}{0.124} & \cellcolor{teal1}{0.058} \\
\midrule
\multirow{4}{*}{\makecell{Average \\Rel. MAE \\over KGs}} 
 & F1 [0.92, 1.00) & \cellcolor{teal9}{0.469} & \cellcolor{teal3}{0.140} & \cellcolor{teal1}{0.0601} & \cellcolor{teal1}{0.022} & \cellcolor{teal1}{0.012} & \cellcolor{teal1}{0.037} & \cellcolor{teal2}{0.110} \\
 & F1 [0.84, 0.92) & \cellcolor{teal9}{0.727} & \cellcolor{teal5}{0.269} & \cellcolor{teal2}{0.106} & \cellcolor{teal1}{0.038} & \cellcolor{teal1}{0.050} & \cellcolor{teal2}{0.085} & \cellcolor{teal2}{0.117} \\
 & F1 [0.76, 0.84) & \cellcolor{teal9}{1.808} & \cellcolor{teal9}{0.734} & \cellcolor{teal4}{0.193} & \cellcolor{teal2}{0.087} & \cellcolor{teal2}{0.074} & \cellcolor{teal2}{0.106} & \cellcolor{teal3}{0.147} \\
 & F1 [0.40, 0.76) & \cellcolor{teal9}{2.699} & \cellcolor{teal9}{1.340} & \cellcolor{teal6}{0.348} & \cellcolor{teal3}{0.167} & \cellcolor{teal2}{0.125} & \cellcolor{teal3}{0.149} & \cellcolor{teal4}{0.200} \\
\bottomrule
\end{tabular}
\caption{\textbf{{\colorbox{clustering}{Tightness of connections.}}} Comparison of relative bias of metrics that capture analyses of tightness of connections, and  ground truth over KGs from the three scanned books. Cell shading intensity corresponds to the magnitude of the value, with darker colors indicating larger magnitudes. Brown is used for negative values, and green for positive values.}
\label{tab:clustering-metrics-comparison}
\end{table}

\begin{table}[ht]
\centering
\begin{tabular}{
  l
  >{\raggedright}p{2cm}c c c cc c
}
%\toprule
 & & \multicolumn{2}{c}{Partition $V^{indiv}$} & \multicolumn{2}{c}{Partition $V^{club}$} & \multicolumn{2}{c}{Partition $V^{club}$} \\
\cmidrule(lr){3-4} \cmidrule(lr){5-6} \cmidrule(lr){7-8}
 & & {\makecell{Degree \\Mean}} & {\makecell{Degree \\Std Dev}} & {\makecell{Degree \\Mean}} & {\makecell{Degree \\Std Dev}} & {Top 10} & {All} \\
\midrule
\multirow{3}{*}{\makecell{Ground\\ Truth}} 
 & Denver & {3.777} & {2.283} & {5.452} & {19.230} & 0 & 0 \\
 & South US (avg) & {2.106} & {1.391} & {4.085} & {8.984} & 0 & 0 \\
 & Rhodesia & {2.631} & {1.516} & {3.081} & {13.140} & 0 & 0 \\
\midrule
\multirow{4}{*}{\makecell{Average Rel.\\ Bias over KGs}} 
 & F1 [0.92, 1.00) & \cellcolor{brown1}{-0.009} & \cellcolor{teal1}{0.001} & \cellcolor{brown1}{-0.018} & \cellcolor{brown1}{-0.011} & n/a & n/a \\
 & F1 [0.84, 0.92) & \cellcolor{brown1}{-0.036} & \cellcolor{brown1}{-0.025} & \cellcolor{brown1}{-0.020} & \cellcolor{brown1}{-0.034} & n/a & n/a \\
 & F1 [0.76, 0.84) & \cellcolor{brown1}{-0.045} & \cellcolor{brown1}{-0.006} & \cellcolor{brown2}{-0.063} & \cellcolor{brown2}{-0.068} & n/a & n/a \\
 & F1 [0.40, 0.76) & \cellcolor{brown2}{-0.067} & \cellcolor{brown1}{-0.017} & \cellcolor{brown3}{-0.132} & \cellcolor{brown3}{-0.156} & n/a & n/a \\
\midrule
\multirow{4}{*}{\makecell{Average Rel.\\ MAE over KGs}} 
 & F1 [0.92, 1.00) & \cellcolor{teal1}{0.014} & \cellcolor{teal1}{0.023} & \cellcolor{teal1}{0.054} & \cellcolor{teal1}{0.041} & 0.041* & 0.151* \\
 & F1 [0.84, 0.92) & \cellcolor{teal1}{0.055} & \cellcolor{teal1}{0.057} & \cellcolor{teal1}{0.102} & \cellcolor{teal2}{0.100} & 0.124* & 0.220* \\
 & F1 [0.76, 0.84) & \cellcolor{teal2}{0.077} & \cellcolor{teal2}{0.069} & \cellcolor{teal4}{0.195} & \cellcolor{teal4}{0.198} & 0.193* & 0.254* \\
 & F1 [0.40, 0.76) & \cellcolor{teal2}{0.092} & \cellcolor{teal2}{0.077} & \cellcolor{teal5}{0.270} & \cellcolor{teal5}{0.311} & 0.277* & 0.259* \\
\bottomrule
\end{tabular}
\caption{\textbf{{\colorbox{degree}{Degree Distribution.}}} Comparison of relative bias of metrics that capture degree distributions over each partition, and  ground truth over KGs from the three scanned books. Cell shading intensity corresponds to the magnitude of the value, with darker colors indicating larger magnitudes. Brown is used for negative values, and green for positive values.}
\label{tab:degree-metrics-comparison}
\end{table}

Tables~\ref{tab:projection-metrics-comparison}, \ref{tab:clustering-metrics-comparison}, and \ref{tab:degree-metrics-comparison} report ground truth metric values for Denver, Southern U.S. states, and Rhodesia datasets. To assess the direction and magnitude of estimation error, we compute the average relative bias and relative MAE over all datasets, across F1 performance ranges: [0.92,1.00), [0.84,0.92), [0.76,0.84), and [0.4,0.76), with the last bin broader due to fewer low-performing model combinations. For the Southern U.S. states, we present averages across all 13 graphs corresponding to each state; notably, the effect of extraction error on downstream graph analyses exhibited very similar patterns in each state.

\begin{figure}
    \centering
    \includegraphics[scale=0.4]{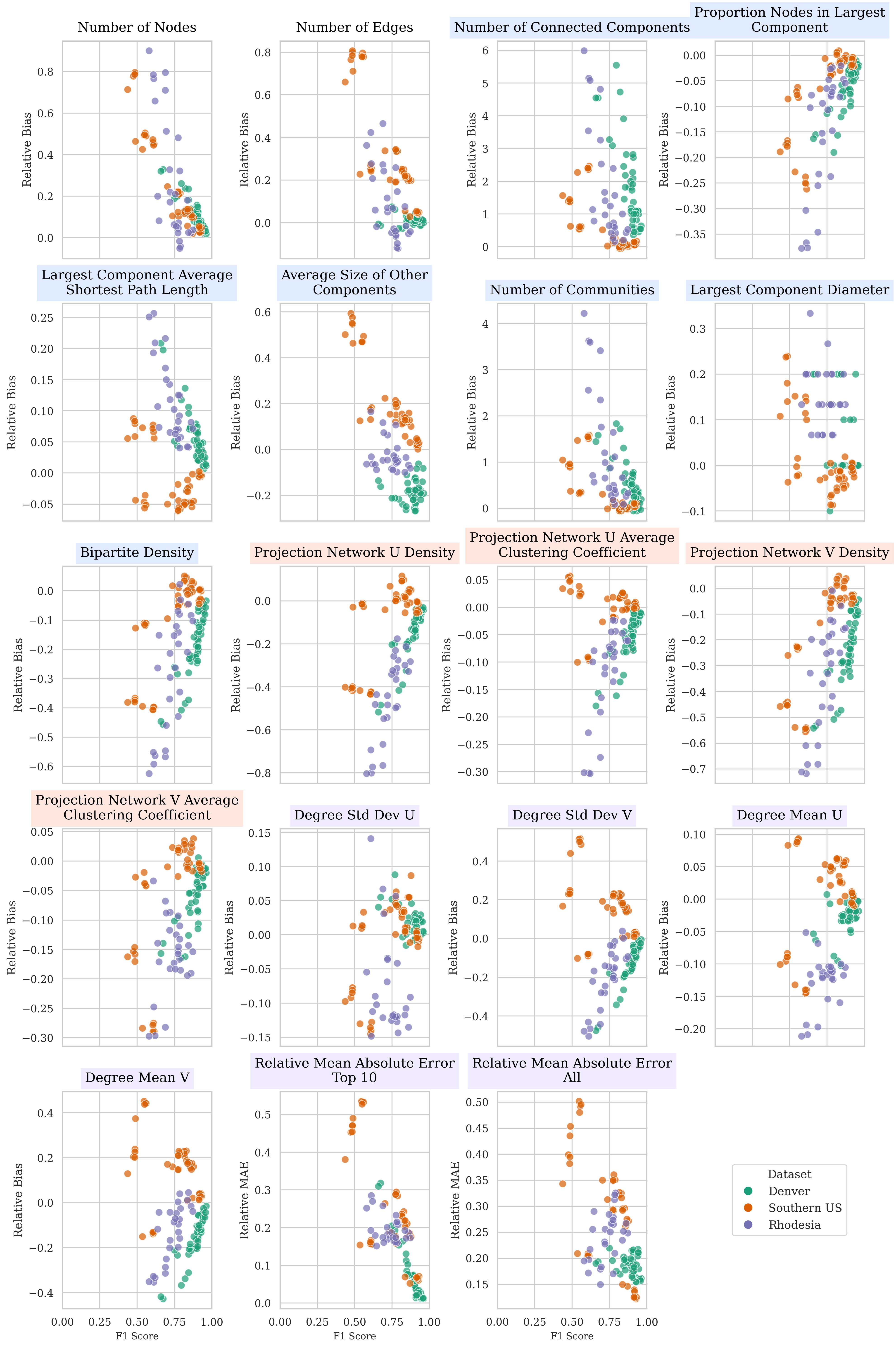}
    \caption{Scatterplots of relative MAE or bias of graph analysis metrics for KG extraction methods under different performance levels.).
    % on extracting articles about randomly selected disaster events of different classes.
    }
    \label{fig:graph-analysis-scatterplots}
\end{figure}

The metrics are grouped by type: {\colorbox{clustering}{tightness of connections}} (Table~\ref{tab:clustering-metrics-comparison}), {\colorbox{projection}{projection networks}} (Table~\ref{tab:projection-metrics-comparison}), and {\colorbox{degree}{degree distributions}} (Table~\ref{tab:degree-metrics-comparison}). Fig.~\ref{fig:graph-analysis-scatterplots} shows individual bias values per dataset and performance level. While the individual values may be noisy, two common patterns emerge:

\vspace{0.3em}
\noindent \textbf{Threshold for small relative MAE.} As shown in Tables~\ref{tab:projection-metrics-comparison}, \ref{tab:clustering-metrics-comparison},  and \ref{tab:degree-metrics-comparison}, and Fig.~\ref{fig:graph-analysis-scatterplots}, when the F1 score exceeds 0.92, the average relative MAE for most metrics stays below 0.08. Exceptions are the numbers of connected components and communities, which show slightly higher bias. %This suggests that an F1 score above $0.92$ may reliably indicate low error in downstream analyses, although the exact threshold may vary due to the binning choice.

\vspace{0.3em}
\noindent\textbf{Consistent over- and underestimation patterns.} Despite some noise in individual bias values, visible in the spread of points in Fig.~\ref{fig:graph-analysis-scatterplots}, there are clear and reproducible aggregate patterns in whether metrics are over- or underestimated, both before (Fig.~\ref{fig:graph-analysis-scatterplots}) and after (Tables~\ref{tab:projection-metrics-comparison}, \ref{tab:clustering-metrics-comparison}, \ref{tab:degree-metrics-comparison}) averaging across graphs and performance ranges. The close alignment between MAE and relative bias further suggests errors consistently skew in one direction rather than varying randomly.

\vspace{0.3em}
\noindent \textbf{Metrics with lower sensitivity to extraction errors.} As reflected by the lighter cell shading in the tables, which corresponds to smaller error magnitudes, certain graph analysis metrics have relatively lower sensitivity to graph extraction errors. 

\vspace{0.3em}
\noindent\textit{\colorbox{clustering}{Connections in extracted bipartite networks are less tight.}} For metrics assessing network connectivity, pronounced trends of over- or underestimation often emerge as performance decreases,
indicating less tightly connected networks, though the magnitude of these effects can vary and a few individual results may deviate from the overall pattern. In particular, the number of connected components and communities increases dramatically as performance decreases, indicating a more fragmented network. Bipartite density consistently drops---even as edge counts may rise---because node counts increase more rapidly, reducing edge-to-possible-edge ratios. The proportion of nodes in the largest component decreases, while average shortest path length and diameter for the largest connected component tend to rise (albeit with some noise), showing a looser and more dispersed network as performance decreases. The average size of smaller components (excluding the largest) also increases, though this trend varies across performance levels.

\vspace{0.3em}
\noindent\textit{\colorbox{projection}{Connections in the projection network of extracted graphs are less tight.}} In projection network analyses, connections consistently become less tight as performance decreases, though some noise is present at the individual graph level. Densities in both the comembership network (projection network $V^{indiv}$) and the organizations-sharing-members network (projection network $V^{club}$) are increasingly underestimated across all datasets and performance levels, both before and after averaging metrics over graphs by performance. Clustering coefficients for the organizations-sharing-members network also show consistent underestimation with declining performance, while clustering in the comembership network declines for two datasets but remains less consistent for the Southern U.S. dataset.

\begin{figure}
    \centering
    \includegraphics[scale=0.28]{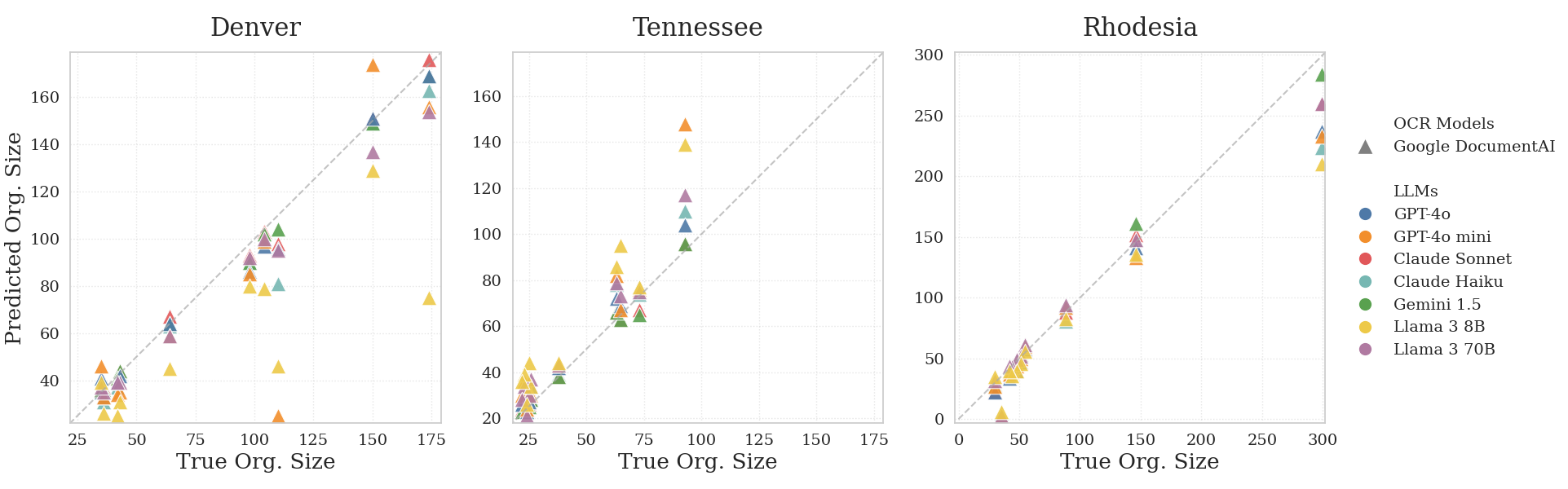}
    \caption{Predicted vs. true sizes for the 10 largest clubs across 7 LLMs on TextRact OCR-outputted text.
    % on extracting articles about randomly selected disaster events of different classes.
    }
    \label{fig:org-perf}
\end{figure}

\vspace{0.3em}
\noindent\textit{\colorbox{degree}{Degree distribution changes show no clear pattern.}}
Degree distribution analyses reveal no consistent trend across datasets. While summary statistics may hint at patterns, scatterplots show no reliable relationship between extraction performance and degree metrics. Comparing the top ten node degrees in ground truth and extracted graphs also shows no clear pattern of over- or underestimation, highlighting the complexity of degree-related errors.

\vspace{0.3em}
\noindent\textbf{Key insights for practitioners.}
Practitioners can draw three main insights to guide future network extraction and analysis efforts, while keeping in mind that individual results may exhibit variability due to inherent noise:

\vspace{0.3em}
\noindent\textit{Performance Thresholds for Reliable Analyses.} We find that when the F1 score exceeds approximately $0.92$, relative MAE for network metrics generally remains below $0.08$, offering a useful guideline for selecting extraction methods likely to yield reliable downstream results. Future work could refine this with more granular performance metrics and by validating these findings on new datasets as more ground truth KGs become available.

\vspace{0.3em}
\noindent\textit{Directional Trends in Error Effects.} We observe consistent directional trends in how errors bias many different analysis metrics, helping practitioners anticipate whether a metric will be over- or underestimated as performance drops. While exact bias magnitude needs further study, and while not all analyses exhibit this behavior, these recurring directional patterns provide useful qualitative guidance. Achieving more precise bias estimates requires further experimentation and simulations using ground truth data---tasks we begin to explore in Section~\ref{sec:sim}, but leave for future research due to their complexity and scope.

\vspace{0.3em}
\noindent \textit{Metrics with Lower Sensitivity to Extraction Error.} Our analysis reveals that certain graph analysis metrics are consistently less affected by extraction errors, exhibiting only small changes as extraction F1 score decreases. We recommend prioritizing these more stable metrics for downstream applications, especially when extraction accuracy is not expected to be high. %For practitioners, this indicates that some network properties can be estimated robustly even when automated KG extraction is imperfect, supporting meaningful analysis. 

\section{Simulation-Based Error Models: Limitations and Comparison with Real Extraction Errors}
\label{sec:sim}
Consistent over- and underestimation trends in bias across performance levels suggest potential predictable patterns in how KG extraction errors affect graph analyses. To test whether these patterns could be replicated without resource-intensive ground truth experiments, we conducted simulations using error models from existing literature. Successfully reproducing these trends in simulation would be an important first step toward constructing fully simulated frameworks for evaluating the impact of KG extraction errors, reducing the reliance on manual ground truth construction.

\vspace{0.3em}
\noindent \textbf{Prior simulation studies on graph errors.} Most theoretical and simulation studies on error effects in downstream graph analysis do not consider automated text-based KG extraction, but rather address extraction errors in general \citep{erman2015effects,lewbel2024ignoring} using simple models such as random or preferential attachment errors \citep{deron2022quantifying,almquist2012random}. Simulations have been developed to better understand downstream effect of graph error in other domains, such as signal processing \citep{miettinen2021modelling,miettinen2018} and protein-protein interactions \citep{chiang2007coverage}.

\vspace{0.3em}
\noindent \textbf{Empirical studies of error sources in text-based extraction.} Although the effect of KG extraction errors from text on downstream graph analyses is not well studied, numerous works have empirically examined error sources in text-based information extraction tasks. For example, \citet{valls2016error} examines errors in extracting information from narratives, stemming from challenges such as coreference resolution, character identification, and role assignment in narrative extraction. \citet{grundkiewicz2013auto} identifies errors related to spelling, grammar, syntax, semantics, and style in text edition histories. In speech-to-text systems, \citet{palmer-ostendorf-2001-improving} finds that errors often arise from out-of-vocabulary words or misrecognition of in-vocabulary terms. Similarly, \citet{das-etal-2022-automatic} categorizes errors in template extraction based on the role of extracted information, which is a much more complex task than extracting affiliation graphs, which are the focus of our study.

\vspace{0.3em}
\noindent \textbf{Simulation framework and error model implementation.} 
We adapted both widely used error models and tailored ones reflecting specific error types observed in Section~\ref{sec:micro}, to simulate edge- and node-level errors at the edge precision and recall levels matching our LLM experiments. Each model simulates only a single type of false positive edge error (e.g., preferential attachment, spelling mistakes), consistent with the simulations used in most prior work. However, these models do not account for the combination and interaction of multiple error types that are prevalent in real-world KG extraction and reflected in our Section~\ref{sec:micro} results; accurately modeling such interactions would require additional parameters and a more complex design. False negatives are generally modeled as random, which is standard in the literature and seems consistent with our findings; manual inspection revealed no clear pattern beyond false negatives caused by the specific error types in Table~\ref{tab:error_distribution}.

For each simulation, we generated graphs $G' = (V^{indiv\prime}, V^{club\prime}, E')$ by retaining a proportion of true edges ($E_{keep}$) and adding false edges ($E_{add}$) to achieve target precision and recall rates. We chose to parameterize our error model using precision and recall because these metrics are widely used in the evaluation of relation and KG extraction methods and directly model the rates observed from real extraction systems such as OCR, relation extraction, or LLM-based pipelines. This approach differs from other work in the literature, which often parameterizes error models with quantities such as the false positive rate.

We define our notation as follows:
\[
E' = E_{keep} \cup E_{add},
\]
where $|E_{keep}| = \lfloor \text{Recall} \cdot |E| \rfloor$ is the number of true edges retained, and
\[
|E_{add}| = \left\lfloor \left(\frac{1}{\text{Precision}} - 1\right) |E_{keep}| \right\rfloor.
\]
We note that this formula ensures that the final edge set $E'$ achieves the specified precision and recall, since $\text{Precision} = |E_{keep}|/(|E_{keep}| + |E_{add}|)$. Intuitively, the term $(1/\text{Precision} - 1)$ is equal to the ratio of false to true positives in $E'$; that is, $(1/\text{Precision} - 1) = (TP + FP)/TP - 1 = FP/TP$, where $TP = |E_{keep}|$ and $FP = |E_{add}|$. Multiplying this ratio by the number of true positives gives the number of false positive edges to add, helping us directly control the error rates to match realistic ones that are observed in automated KG extraction pipelines.

%\noindent 
We applied a variety of error models to introduce perturbations in both edge sets (i.e., modifying edge presence or absence) and node sets (i.e., adding or removing nodes) \citep{wang2012measurement}. These models control the specific types of errors introduced, such as false positive and false negative edges or spurious and missing nodes, while the targeted precision and recall rates determine the magnitude of these perturbations. The error models include both standard approaches from the literature and models that are designed to simulate node and edge modifications resulting from misspellings or mistaken entities, reflecting common error sources in automated KG extraction as identified by our manual analysis in Section~\ref{sec:micro}.

\vspace{0.3em}
\noindent \textbf{Error models focusing on edge errors.}
These are widely used models that implement edge addition and removal based on precision and recall, while keeping the node set fixed.

%\vspace{0.3em}
\begin{itemize}
%In this model, edges are randomly removed with a probability of $\frac{1}{|E|}$ and edges are randomly added with a probability of $\frac{1}{|V^{indiv}| |V^{club}|}$ \citep{deron2022quantifying,almquist2012random}. This approach aligns with simple random error assumptions that are commonly used in network error modeling. %, and the random removal of edges to match the recall rate is consistent with our observations of retained edges in ground truth graphs (Section~\ref{sec:micro}) and with error models in the literature.
\item\textit{Random edge removal and random edge addition.} Edges are removed uniformly at random resulting in $E_{\text{keep}}$, and new edges are added by randomly sampling pairs of nodes from the two bipartite sets uniformly from all possible unconnected pairs, forming $E_{\text{add}}$. The probability of adding an edge between any pair of nodes $v^{indiv}_i$ and $v^{club}_k$ is

\[
P\big((v^{indiv}_i, v^{club}_k)\big) =
\begin{cases}
\frac{1}{|V^{indiv}| \cdot |V^{club}| - |E|} & \text{if } (v^{indiv}_i, v^{club}_k) \notin E \\
0 \quad \text{otherwise}
\end{cases}
\] This approach models the simplest case of random error, which is consistent with standard assumptions in network error modeling \citep{deron2022quantifying,almquist2012random}.
\item  \textit{Random edge removal and preferential attachment edge addition.} 
Edges are removed uniformly at random resulting in $E_{\text{keep}}$ as above, but new edges for $E_{\text{add}}$ are added by independently sampling pairs of nodes, one from each bipartite set, with the probability of selecting each node proportional to its degree in the original graph (preferential attachment) \citep{lin2005are}. Specifically, the probability of adding a new edge between nodes \(v^{indiv}_i\) and \(v^{club}_k\) is:
\[
P\big((v^{indiv}_i, v^{club}_k)\big) = 
\frac{\text{deg}(v^{indiv}_i)}{\sum_{j} \text{deg}(v^{indiv}_j)}
\cdot
\frac{\text{deg}(v^{club}_k)}{\sum_{l} \text{deg}(v^{club}_l)}
\]
This approach biases edge addition toward higher-degree nodes in each bipartite set, therefore preserving the original degree distributions. 
\end{itemize}

 %The random removal of edges is consistent with our observations of ground truth retained edges in Section xx and the edge addition strategy is consistent with models in the literature. ...however not the no addition of nodes

\vspace{0.3em}
\noindent \textbf{Error models focusing on node-level errors.} To better reflect the frequent node-level errors observed in real-world KG extraction, we implemented models that modify the node set itself by adding or splitting nodes \citep{deron2022quantifying, almquist2012random, nakajima2021measure}. This is consistent with the spelling errors and node overestimation phenomena identified in Section~\ref{sec:micro}.

\begin{itemize}
\item \textit{Disaggregating nodes to simulate spelling errors.} We define a node splitting error model for bipartite graphs to simulate the impact of spelling errors in entity recognition. This model redirects a subset of original edges to new ``misspelled" nodes.\\

We first randomly select $|E_{\text{not misspelled}}| = \lfloor \max(Precision, Recall) \cdot |E| \rfloor$ edges from $E$ to retain between original node pairs. This choice prevents over-perturbation in the simulated error process, because it ensures that both target precision and recall rates can still be satisfied after edge redirection. For each remaining edge in $E_{\text{misspelled}} = E-E_{\text{not mispelled}}$, we simulate a misspelling by randomly selecting one endpoint (with equal probability) to keep as the correct node, and replacing the other endpoint with a newly created node in the opposite partition.\\

After this edge redirection step, if the number of true edges from the original graph is higher than $|E_{keep}|$, additional true positive edges are randomly removed until $|E_{keep}|$, matching the specified recall, is achieved. To meet the target precision, we further introduce false positive edges by connecting randomly selected original nodes to new nodes in the opposite partition until $|E_{add}|$ is achieved.\\

This procedure ensures that the perturbed edge set matches the specified precision and recall rates, providing a realistic simulation of entity splitting and name variation errors in downstream graph analysis.

\item \textit{Introducing false positive nodes to simulate misidentified entities.} We simulate spurious entities by adding $|E_{add}|$ edges between existing nodes and newly created nodes, as per the target precision rate, to reflect cases where non-existent entities are introduced into the graph. Specifically, for each additional false positive edge, we randomly select one of the two bipartite partitions with equal probability ($0.5$), sample a random node from that partition, and connect it to a newly created node in the opposite partition. To match the target recall, we also randomly remove $|E_{keep}|$ edges.

\end{itemize}

\noindent As in prior work, these single error type models do not account for the complex, interacting error types that we found to occur in real-world KG extraction (Section~\ref{sec:micro}).

\begin{table}[ht]
\centering
\begin{tabular}{ccccccccc}
%\toprule
 & & \multicolumn{3}{c}{\makecell{Full network}} & \multicolumn{3}{c}{\makecell{Largest Conn. Comp.}} & {\makecell{Rest \\Comp.}}\\
\cmidrule(lr){3-5} \cmidrule(lr){6-8} \cmidrule(lr){9-9}
 & & {\makecell{\# Conn.\\ Comp.}} & {\makecell{\# Com-\\munity}} & {\makecell{Bipart. \\Density}} & {\makecell{Prop. of\\ Nodes}} & {\makecell{Avg Short.\\ Path Len.}} & {Diam.} & {\makecell{Avg \\Size}} \\
\midrule
\multirow{4}{*}{\makecell{\textit{Random}\\ \textit{Error}} }
 & F1 [0.92, 1.00) & \cellcolor{teal3}0.182 & \cellcolor{brown1}-0.039 & \cellcolor{teal1}0.024 & \cellcolor{teal1}0.036 & \cellcolor{brown1}-0.005 & \cellcolor{brown1}-0.006 & \cellcolor{brown7}-0.397 \\
 & F1 [0.84, 0.92) & \cellcolor{teal6}0.369 & \cellcolor{teal2}0.094 & \cellcolor{teal1}0.036 & \cellcolor{teal1}0.046 & \cellcolor{teal1}0.004 & \cellcolor{brown1}-0.022 & \cellcolor{brown9}-0.513 \\
 & F1 [0.76, 0.84) & \cellcolor{teal8}0.462 & \cellcolor{teal2}0.109 & \cellcolor{teal2}0.081 & \cellcolor{teal1}0.056 & \cellcolor{brown1}-0.012 & \cellcolor{brown2}-0.066 & \cellcolor{brown9}-0.582 \\
 & F1 [0.40, 0.76) & \cellcolor{teal1}0.037 & \cellcolor{brown3}-0.171 & \cellcolor{teal4}0.246 & \cellcolor{teal2}0.082 & \cellcolor{brown1}-0.045 & \cellcolor{brown3}-0.152 & \cellcolor{brown9}-0.634 \\
\midrule
\multirow{4}{*}{\makecell{\textit{Preferential}\\ \textit{Attachment}\\ \textit{Error}} }
 & F1 [0.92, 1.00) & \cellcolor{teal9}0.690 & \cellcolor{teal4}0.202 & \cellcolor{teal1}0.014 & \cellcolor{teal1}0.004 & \cellcolor{brown1}-0.047 & \cellcolor{brown2}-0.081 & \cellcolor{brown6}-0.351 \\
 & F1 [0.84, 0.92) & \cellcolor{teal9}1.300 & \cellcolor{teal9}0.579 & \cellcolor{teal1}0.018 & \cellcolor{brown1}-0.008 & \cellcolor{brown2}-0.063 & \cellcolor{brown2}-0.094 & \cellcolor{brown8}-0.478 \\
 & F1 [0.76, 0.84) & \cellcolor{teal9}2.291 & \cellcolor{teal9}0.951 & \cellcolor{teal1}0.050 & \cellcolor{brown1}-0.022 & \cellcolor{brown2}-0.100 & \cellcolor{brown3}-0.145 & \cellcolor{brown9}-0.562 \\
 & F1 [0.40, 0.76) & \cellcolor{teal9}4.189 & \cellcolor{teal9}1.795 & \cellcolor{teal3}0.180 & \cellcolor{brown2}-0.073 & \cellcolor{brown3}-0.169 & \cellcolor{brown4}-0.207 & \cellcolor{brown9}-0.626 \\
\midrule
\multirow{4}{*}{\makecell{\textit{Node} \\\textit{Addition}\\ \textit{Error}}} 
 & F1 [0.92, 1.00) & \cellcolor{teal9}0.959 & \cellcolor{teal8}0.479 & \cellcolor{brown4}-0.211 & \cellcolor{brown1}-0.033 & \cellcolor{teal2}0.070 & \cellcolor{teal2}0.082 & \cellcolor{brown4}-0.193 \\
 & F1 [0.84, 0.92) & \cellcolor{teal9}1.713 & \cellcolor{teal9}1.038 & \cellcolor{brown6}-0.329 & \cellcolor{brown2}-0.067 & \cellcolor{teal3}0.125 & \cellcolor{teal3}0.161 & \cellcolor{brown4}-0.226 \\
 & F1 [0.76, 0.84) & \cellcolor{teal9}3.001 & \cellcolor{teal9}1.716 & \cellcolor{brown8}-0.468 & \cellcolor{brown2}-0.114 & \cellcolor{teal4}0.202 & \cellcolor{teal4}0.231 & \cellcolor{brown4}-0.188 \\
 & F1 [0.40, 0.76) & \cellcolor{teal9}5.036 & \cellcolor{teal9}2.942 & \cellcolor{brown9}-0.627 & \cellcolor{brown4}-0.203 & \cellcolor{teal6}0.336 & \cellcolor{teal7}0.377 & \cellcolor{brown2}-0.069 \\
\midrule
\multirow{4}{*}{\makecell{\textit{Node}\\ \textit{Disaggregation}\\ \textit{(Misspelling)} \\\textit{Error}}}
 & F1 [0.92, 1.00) & \cellcolor{teal9}2.037 & \cellcolor{teal9}0.802 & \cellcolor{brown4}-0.213 & \cellcolor{brown2}-0.064 & \cellcolor{teal1}0.042 & \cellcolor{teal1}0.040 & \cellcolor{brown6}-0.347 \\
 & F1 [0.84, 0.92) & \cellcolor{teal9}4.009 & \cellcolor{teal9}1.923 & \cellcolor{brown6}-0.341 & \cellcolor{brown3}-0.142 & \cellcolor{teal2}0.075 & \cellcolor{teal2}0.064 & \cellcolor{brown7}-0.413 \\
 & F1 [0.76, 0.84) & \cellcolor{teal9}7.890 & \cellcolor{teal9}3.513 & \cellcolor{brown8}-0.475 & \cellcolor{brown4}-0.232 & \cellcolor{teal2}0.116 & \cellcolor{teal3}0.143 & \cellcolor{brown8}-0.458 \\
 & F1 [0.40, 0.76) & \cellcolor{teal9}15.390 & \cellcolor{teal9}6.992 & \cellcolor{brown9}-0.660 & \cellcolor{brown7}-0.423 & \cellcolor{teal3}0.182 & \cellcolor{teal4}0.193 & \cellcolor{brown8}-0.495 \\
\midrule
\multirow{4}{*}{\makecell{Real Error}} 
 & F1 [0.92, 1.00) & \cellcolor{teal8}0.467 & \cellcolor{teal3}0.126 & \cellcolor{brown1}-0.051 & \cellcolor{brown1}-0.022 & \cellcolor{teal1}0.0109 & \cellcolor{teal1}0.005 & \cellcolor{brown2}-0.068 \\
 & F1 [0.84, 0.92) & \cellcolor{teal9}0.716 & \cellcolor{teal5}0.256 & \cellcolor{brown2}-0.090 & \cellcolor{brown1}-0.037 & \cellcolor{teal1}0.032 & \cellcolor{teal1}0.051 & \cellcolor{brown1}-0.055 \\
 & F1 [0.76, 0.84) & \cellcolor{teal9}1.792 & \cellcolor{teal9}0.717 & \cellcolor{brown3}-0.172 & \cellcolor{brown2}-0.084 & \cellcolor{teal1}0.043 & \cellcolor{teal1}0.042 & \cellcolor{brown1}-0.034 \\
 & F1 [0.40, 0.76) & \cellcolor{teal9}2.697 & \cellcolor{teal9}1.337 & \cellcolor{brown6}-0.345 & \cellcolor{brown3}-0.167 & \cellcolor{teal2}0.107 & \cellcolor{teal2}0.124 & \cellcolor{teal1}0.058 \\
\bottomrule
\end{tabular}
\caption{\textbf{{\colorbox{clustering}{Tightness of connections metrics.}}}Comparison of relative bias in network structure metrics across different error types and F1 score ranges. Cell shading intensity corresponds to the magnitude of the value, with darker colors indicating larger magnitudes. Brown is used for negative values, and green for positive values.}
\label{tab:clustering-structure-metrics-error}
\end{table}

\begin{table}[ht]
\centering

\begin{tabular}{>{\raggedright}p{2.2cm}>{\raggedright}p{2cm}cccc}
%\toprule
 & & \multicolumn{2}{c}{\makecell{Comembership Network\\ (Projection Network $V^{indiv}$)}} & \multicolumn{2}{c}{\makecell{Organizations Sharing\\ Members (Projection Network $V^{club}$)}} \\
\cmidrule(lr){3-4} \cmidrule(lr){5-6}
 & & Density & \makecell{Avg Clustering\\ Coefficient} & Density & \makecell{Avg Clustering\\ Coefficient} \\
\midrule
\multirow{4}{*}{\makecell{\textit{Random Error}}} 
 & F1 [0.92, 1.00) & \cellcolor{brown1}-0.052 & \cellcolor{brown1}-0.023 & \cellcolor{teal3}0.153 & \cellcolor{brown1}-0.058 \\
 & F1 [0.84, 0.92) & \cellcolor{brown3}-0.144 & \cellcolor{brown1}-0.057 & \cellcolor{teal4}0.237 & \cellcolor{brown2}-0.120 \\
 & F1 [0.76, 0.84) & \cellcolor{brown4}-0.202 & \cellcolor{brown2}-0.087 & \cellcolor{teal8}0.456 & \cellcolor{brown3}-0.181 \\
 & F1 [0.40, 0.76) & \cellcolor{brown5}-0.274 & \cellcolor{brown3}-0.166 & \cellcolor{teal9}1.166 & \cellcolor{brown5}-0.257 \\
\midrule
\multirow{4}{*}{\makecell{\textit{Preferential}\\ \textit{Attachment Error}}} 
 & F1 [0.92, 1.00) & \cellcolor{teal1}0.030 & \cellcolor{brown1}-0.023 & \cellcolor{teal3}0.125 & \cellcolor{brown1}-0.025 \\
 & F1 [0.84, 0.92) & \cellcolor{teal1}0.017 & \cellcolor{brown1}-0.048 & \cellcolor{teal4}0.213 & \cellcolor{brown2}-0.077 \\
 & F1 [0.76, 0.84) & \cellcolor{teal2}0.079 & \cellcolor{brown2}-0.071 & \cellcolor{teal7}0.413 & \cellcolor{brown2}-0.110 \\
 & F1 [0.40, 0.76) & \cellcolor{teal4}0.249 & \cellcolor{brown2}-0.112 & \cellcolor{teal9}1.055 & \cellcolor{brown3}-0.174 \\
\midrule
\multirow{4}{*}{\makecell{\textit{Node Addition}\\ \textit{Error}}} 
 & F1 [0.92, 1.00) & \cellcolor{brown4}-0.238 & \cellcolor{brown2}-0.070 & \cellcolor{brown4}-0.239 & \cellcolor{brown1}-0.045 \\
 & F1 [0.84, 0.92) & \cellcolor{brown7}-0.409 & \cellcolor{brown2}-0.115 & \cellcolor{brown7}-0.386 & \cellcolor{brown2}-0.092 \\
 & F1 [0.76, 0.84) & \cellcolor{brown9}-0.577 & \cellcolor{brown3}-0.169 & \cellcolor{brown9}-0.542 & \cellcolor{brown3}-0.148 \\
 & F1 [0.40, 0.76) & \cellcolor{brown9}-0.771 & \cellcolor{brown4}-0.232 & \cellcolor{brown9}-0.700 & \cellcolor{brown4}-0.215 \\
\midrule
\multirow{4}{*}{\makecell{\textit{Node}\\ \textit{Disaggregation}\\ \textit{(Misspelling)} \\\textit{Error}}} 
 & F1 [0.92, 1.00) & \cellcolor{brown5}-0.258 & \cellcolor{brown2}-0.105 & \cellcolor{brown5}-0.308 & \cellcolor{brown3}-0.178 \\
 & F1 [0.84, 0.92) & \cellcolor{brown8}-0.455 & \cellcolor{brown4}-0.198 & \cellcolor{brown8}-0.476 & \cellcolor{brown5}-0.304 \\
 & F1 [0.76, 0.84) & \cellcolor{brown9}-0.610 & \cellcolor{brown5}-0.297 & \cellcolor{brown9}-0.636 & \cellcolor{brown8}-0.451 \\
 & F1 [0.40, 0.76) & \cellcolor{brown9}-0.822 & \cellcolor{brown8}-0.480 & \cellcolor{brown9}-0.830 & \cellcolor{brown9}-0.660 \\
\midrule
\multirow{4}{*}{\makecell{Real Error}}
 & F1 [0.92, 1.00) & \cellcolor{brown1}-0.046 & \cellcolor{brown1}-0.008 & \cellcolor{brown2}-0.074 & \cellcolor{brown1}-0.016 \\
 & F1 [0.84, 0.92) & \cellcolor{brown3}-0.136 & \cellcolor{brown1}-0.032 & \cellcolor{brown3}-0.132 & \cellcolor{brown2}-0.070 \\
 & F1 [0.76, 0.84) & \cellcolor{brown4}-0.205 & \cellcolor{brown1}-0.062 & \cellcolor{brown4}-0.227 & \cellcolor{brown2}-0.090 \\
 & F1 [0.40, 0.76) & \cellcolor{brown7}-0.401 & \cellcolor{brown2}-0.107 & \cellcolor{brown7}-0.429 & \cellcolor{brown3}-0.143 \\
\bottomrule
\end{tabular}
\caption{\textbf{{\colorbox{projection}{Analyses over Projection Networks.}}} Comparison of relative bias of metrics that capture analyses of projection networks, and ground truth over KGs from the three scanned books. Cell shading intensity corresponds to the magnitude of the value, with darker colors indicating larger magnitudes. Brown is used for negative values, and green for positive values.}
\label{tab:projection-metrics-comparison-error}
\end{table}

\begin{table}[ht]
\centering
\begin{tabular}{>{\raggedright}p{2.2cm}>{\raggedright}p{2cm}cccccc}
%\toprule
 & & \multicolumn{2}{c}{Partition $V^{indiv}$} & \multicolumn{2}{c}{Partition $V^{club}$} & \multicolumn{2}{c}{Partition $V^{club}$} \\
\cmidrule(lr){3-4} \cmidrule(lr){5-6} \cmidrule(lr){7-8}
 & & {\makecell{Degree \\Mean}} & {\makecell{Degree \\Std Dev}} & {\makecell{Degree \\Mean}} & {\makecell{Degree \\Std Dev}} & {Top 10} & {All} \\
\midrule
\multirow{4}{*}{\makecell{\textit{Random Error}} }
 & F1 [0.92, 1.00) & \cellcolor{teal1}0.024 & \cellcolor{teal1}0.014 & \cellcolor{teal1}0.024 & \cellcolor{brown1}-0.049 & \cellcolor{teal1}0.051 & \cellcolor{teal5}0.286 \\
 & F1 [0.84, 0.92) & \cellcolor{teal1}0.036 & \cellcolor{teal1}0.018 & \cellcolor{teal1}0.036 & \cellcolor{brown2}-0.110 & \cellcolor{teal2}0.099 & \cellcolor{teal8}0.483 \\
 & F1 [0.76, 0.84) & \cellcolor{teal2}0.081 & \cellcolor{teal1}0.038 & \cellcolor{teal2}0.081 & \cellcolor{brown3}-0.166 & \cellcolor{teal3}0.162 & \cellcolor{teal9}0.791 \\
 & F1 [0.40, 0.76) & \cellcolor{teal4}0.246 & \cellcolor{teal2}0.096 & \cellcolor{teal4}0.246 & \cellcolor{brown5}-0.271 & \cellcolor{teal4}0.246 & \cellcolor{teal9}1.442 \\
\midrule
\multirow{4}{*}{\makecell{\textit{Preferential}\\ \textit{Attachment Error}} }
 & F1 [0.92, 1.00) & \cellcolor{teal1}0.014 & \cellcolor{teal2}0.064 & \cellcolor{teal1}0.014 & \cellcolor{teal1}0.007 & \cellcolor{teal1}0.051 & \cellcolor{teal2}0.115 \\
 & F1 [0.84, 0.92) & \cellcolor{teal1}0.018 & \cellcolor{teal3}0.130 & \cellcolor{teal1}0.018 & \cellcolor{brown1}-0.008 & \cellcolor{teal2}0.092 & \cellcolor{teal4}0.232 \\
 & F1 [0.76, 0.84) & \cellcolor{teal1}0.050 & \cellcolor{teal4}0.246 & \cellcolor{teal1}0.050 & \cellcolor{teal1}0.013 & \cellcolor{teal3}0.139 & \cellcolor{teal7}0.387 \\
 & F1 [0.40, 0.76) & \cellcolor{teal3}0.180 & \cellcolor{teal9}0.605 & \cellcolor{teal3}0.180 & \cellcolor{teal2}0.105 & \cellcolor{teal4}0.236 & \cellcolor{teal9}0.716 \\
\midrule
\multirow{4}{*}{\makecell{\textit{Node Addition}\\ \textit{Error}} }
 & F1 [0.92, 1.00) & \cellcolor{brown2}-0.076 & \cellcolor{brown1}-0.015 & \cellcolor{brown3}-0.126 & \cellcolor{brown2}-0.117 & \cellcolor{teal1}0.049 & \cellcolor{teal3}0.136 \\
 & F1 [0.84, 0.92) & \cellcolor{brown3}-0.133 & \cellcolor{brown1}-0.042 & \cellcolor{brown4}-0.202 & \cellcolor{brown4}-0.211 & \cellcolor{teal2}0.104 & \cellcolor{teal4}0.200 \\
 & F1 [0.76, 0.84) & \cellcolor{brown4}-0.195 & \cellcolor{brown2}-0.085 & \cellcolor{brown5}-0.290 & \cellcolor{brown6}-0.316 & \cellcolor{teal3}0.166 & \cellcolor{teal5}0.259 \\
 & F1 [0.40, 0.76) & \cellcolor{brown5}-0.268 & \cellcolor{brown3}-0.154 & \cellcolor{brown7}-0.384 & \cellcolor{brown8}-0.470 & \cellcolor{teal5}0.267 & \cellcolor{teal5}0.280 \\
\midrule
\multirow{4}{*}{\makecell{\textit{Node}\\ \textit{Disaggregation}\\ \textit{(Misspelling)} \\\textit{Error}}}
 & F1 [0.92, 1.00) & \cellcolor{brown2}-0.117 & \cellcolor{brown1}-0.045 & \cellcolor{brown3}-0.156 & \cellcolor{brown2}-0.123 & \cellcolor{teal2}0.076 & \cellcolor{teal2}0.068 \\
 & F1 [0.84, 0.92) & \cellcolor{brown4}-0.216 & \cellcolor{brown2}-0.090 & \cellcolor{brown5}-0.253 & \cellcolor{brown4}-0.223 & \cellcolor{teal3}0.155 & \cellcolor{teal2}0.124 \\
 & F1 [0.76, 0.84) & \cellcolor{brown6}-0.320 & \cellcolor{brown3}-0.150 & \cellcolor{brown6}-0.367 & \cellcolor{brown6}-0.323 & \cellcolor{teal4}0.234 & \cellcolor{teal3}0.174 \\
 & F1 [0.40, 0.76) & \cellcolor{brown8}-0.479 & \cellcolor{brown5}-0.272 & \cellcolor{brown9}-0.538 & \cellcolor{brown8}-0.494 & \cellcolor{teal7}0.379 & \cellcolor{teal4}0.240 \\
\midrule
\multirow{4}{*}{\makecell{Real Error}}
 & F1 [0.92, 1.00) & \cellcolor{brown1}-0.009 & \cellcolor{teal1}0.001 & \cellcolor{brown1}-0.018 & \cellcolor{brown1}-0.011 & \cellcolor{teal1}0.041 & \cellcolor{teal3}0.151 \\
 & F1 [0.84, 0.92) & \cellcolor{brown1}-0.036 & \cellcolor{brown1}-0.025 & \cellcolor{brown1}-0.020 & \cellcolor{brown1}-0.034 & \cellcolor{teal2}0.124 & \cellcolor{teal4}0.220 \\
 & F1 [0.76, 0.84) & \cellcolor{brown1}-0.045 & \cellcolor{brown1}-0.006 & \cellcolor{brown2}-0.063 & \cellcolor{brown2}-0.068 & \cellcolor{teal4}0.193 & \cellcolor{teal5}0.254 \\
 & F1 [0.40, 0.76) & \cellcolor{brown2}-0.067 & \cellcolor{brown1}-0.017 & \cellcolor{brown3}-0.132 & \cellcolor{brown3}-0.156 & \cellcolor{teal5}0.277 & \cellcolor{teal5}0.259 \\
\bottomrule
\end{tabular}
\caption{\textbf{{\colorbox{degree}{Degree Distribution Analyses.} }}Comparison of relative bias in degree distribution metrics across different partitions and error types. Cell shading intensity corresponds to the magnitude of the value, with darker colors indicating larger magnitudes. Brown is used for negative values, and green for positive values.}
\label{tab:degree-distribution-metrics-error}
\end{table}

\vspace{0.3em}
\noindent \textbf{Comparison of simulated and real extraction error patterns.} For each simulated graph, we measured the average relative bias of metrics within the same performance ranges as in Section~\ref{sec:macro}.

\noindent \textbf{\textit{Inconsistent bias direction in common error models.}} Comparing the patterns of relative bias from these simulations with those observed in graphs affected by real-world errors, we found that widely used error models in the literature (e.g., random edge error and preferential attachment) fail to replicate the directional trends of bias seen in real-world errors of the same magnitude. For example, in Table~\ref{tab:clustering-structure-metrics-error}, as performance decreases, bipartite density is consistently overestimated under the random and preferential attachment error models but not under real-world errors. Similarly, in Table~\ref{tab:projection-metrics-comparison-error}, projection network density is consistently overestimated under these error models but not under real-world errors.

\vspace{0.3em}
\noindent \textbf{\textit{Inconsistent bias magnitude in simulated errors.}} The node error models that are consistent with error types from our manual analysis do mirror the trend of bias as performance drops, but the magnitude is greatly exaggerated. In Tables~\ref{tab:clustering-structure-metrics-error}, \ref{tab:projection-metrics-comparison-error} and \ref{tab:degree-distribution-metrics-error}, the bias of every analysis stemming from simulated error related to node addition or disaggregation is over four times greater than that of real-world errors. Random and preferential attachment models produce more realistic magnitudes, but their bias direction often contradicts those found in real extraction.

\vspace{0.3em}
\noindent\textbf{Recommendation.} 
These findings show that common error models, such as random or preferential edge perturbation, are inadequate for simulating the actual errors produced by automatic KG extraction from text. As our manual evaluation in Section~\ref{sec:micro} demonstrates, real extraction errors are caused by a diverse mix of sources: misspellings and node splits, redirected edges due to entity confusion, and introduction of spurious nodes, among others. In contrast, many simulation approaches model only one error type at a time and do not capture these interacting errors of multiple types, therefore failing to reproduce the bias or variability seen in downstream graph analyses of automatically extracted KGs. This demonstrates that accurately modeling and predicting these effects is difficult and that considerable progress is still needed. Future work could develop mixed-error models that reflect the full diversity and interactions of real world error sources observed in KG extraction from text. Grounded in manual inspection of automated KG extraction output, these models will be foundational for realistic simulation and robust downstream analysis.

%These findings demonstrate the challenges of using existing error models to simulate real-world KG extraction errors. We recommend developing heterogeneous error models that capture the range and interaction of error types revealed by manual inspection.  %As automatic KG extraction methods improve and become more scalable, understanding their impact on downstream analyses becomes increasingly important. The insights from our study can inform practitioners about the effects of error in single-relation-type settings. For broader applications, we recommend experimenting with new error models consistent with our findings on real-world errors to construct more realistic simulations. Additionally, performing more real-world experiments on other KG and text pairs—if ground-truth KGs become available—would further advance this line of research.

\section{Conclusion}
Understanding the detailed effects of KG extraction errors on downstream graph analyses is increasingly important as automated methods, particularly those using LLMs, make large-scale KG construction from text feasible across diverse domains. Although previously overlooked, the errors introduced by these automated methods pose significant challenges for real-world applications. As practitioners depend more on automated KG extraction for large-scale analyses, understanding how these errors propagate and impact results is vital. Our affiliation graph case study provides important insights into the reliability and accuracy of automated KG extraction and its implications for downstream analysis, and points to the need for more realistic, empirically grounded error modeling to address the complexity of real-world extraction errors. %It also offers guidance for improving error modeling, calling for the development of heterogeneous, empirically grounded models that capture the true complexity of real-world extraction errors.

\begin{appendices}

\section{LLM prompts} 
\label{app:prompts}
We used the same prompts across LLMs for each dataset. The prompt for the Denver dataset was:

\begin{quote}
\textit{[text entry]}\\

\noindent \textit{Extract relations in the format of person-name; member; club from the text, indicating that person-name is a member of club. In the text, — indicates a different club. List each relation as a bullet point. An example is: }\\

\noindent \textit{[example text entry and outputted relation triplets]}\\
\end{quote}

\noindent The prompt for the Southern US dataset was:\\

\begin{quote}
\textit{[text entry]}\\

\noindent \textit{Extract relations in the format of person-name; member; club from the text, indicating that person-name is a member of club. Schools are not clubs. If no title such as 'Mr.', 'Mrs.', 'Miss', 'Rev.', 'Dr.', etc. is in front of a list of clubs, then the first two persons mentioned are members in each of the listed clubs. Otherwise, only the person with the corresponding title is in each of the listed clubs. List each relation as a bullet point.}\\

\noindent \textit{[example text entry and outputted relation triplets]}\\
\end{quote}

\noindent The prompt for the Rhodesia dataset was: \\

\begin{quote}
\textit{[text entry]}\\

\noindent \textit{Extract relations in the format of person-name; member; club from the text, indicating that person-name is a member of club. Consider only the clubs that are listed after ``Club(s):". List each relation as a bullet point. An example is:}\\

\noindent \textit{[example text entry and outputted relation triplets]}
\end{quote}

\section{Details for evaluating triplets}
\label{app:eval}

For evaluation, we compared the triplets $\langle$\textsc{Person}, \textsc{member}, \textsc{Club}$\rangle$ outputted by the LLM to those in the ground truth set. A true positive required that all three components of an LLM triplet match that of a ground truth triplet. However, we allowed more flexible matching for individual \textsc{Person} and \textsc{Club} entities. For \textsc{club} names, LLMs sometimes included additional information, such as relevant parenthetical details, truncated a very long name, or expanded abbreviations. We did not penalize the LLM in these cases because these were spelling, rather than semantic, issues. Therefore, we identified a positive entity match when either the LLM entity matched the ground truth exactly, or when both exceeded ten characters and one was a substring of the other. We also allowed matches between abbreviations and full words or names (e.g., ``Bulawayo" and ``Byo"; ``Assn" and ``Association"). All spaces and punctuation were removed during the comparison. For \textsc{Person} entities, we also required that titles such as ``Mr" or ``Mrs" needed to be correct if present in the text. 

To ensure rigorous evaluation, we enforced a one-to-one correspondence between matched triplets: each LLM-extracted triplet could be paired with at most one ground truth triplet, and vice versa—once a pair was counted as a true positive, both were excluded from further matching. We validated our matching rules by manually reviewing the results of 150 entity pair comparisons for each dataset and confirmed that our approach consistently enforced correct and unique, non-overlapping matches.




\end{appendices}

\bibliographystyle{sn-basic}
\bibliography{references}% common bib file

\end{document}